\documentclass[conference,letterpaper]{IEEEtran}
\usepackage[left=19.1mm, right=19.1mm, top=18.5mm, bottom=19.9mm]{geometry}
\IEEEoverridecommandlockouts
\usepackage{cite}
\DeclareFontFamily{OT1}{pzc}{}
\DeclareFontShape{OT1}{pzc}{m}{it}{<-> s * [1.10] pzcmi7t}{}
\DeclareMathAlphabet{\mathpzc}{OT1}{pzc}{m}{it}
\usepackage{amsmath,amssymb,amsfonts,amsthm}
\usepackage{multicol}

\newtheoremstyle{assumptionstyle}
   {0pt}
   {0pt}
   {}
   {11pt}
   {\em}
   {\em}
   {5pt}
   {}
\theoremstyle{assumptionstyle}

\usepackage{algorithm}
\usepackage{algorithmic}
\usepackage{graphicx}
\usepackage{textcomp}
\usepackage{xcolor}
\def\BibTeX{{\rm B\kern-.05em{\sc i\kern-.025em b}\kern-.08em
    T\kern-.1667em\lower.7ex\hbox{E}\kern-.125emX}}

\usepackage{todonotes}
\usepackage{comment}
\newcommand{\myequation}{\begin{equation}}
\newcommand{\myendequation}{\end{equation}}
\let\[\myequation
\let\]\myendequation
\allowdisplaybreaks

\begin{document}
\bstctlcite{IEEEexample:BSTcontrol}

\title{\vspace*{+6.mm}  Universal Hysteresis Identification Using Extended Preisach Neural Network
\thanks{*Corresponding Author is with the University of Virginia, Department of Engineering Systems and Enviropnment, ffs5da@virginia.edu}
}

\author{\IEEEauthorblockN{M. Farrokh, M. S. Dizaji, F. S. Dizaji, and N. Moradinasab}
}

\maketitle

\begin{abstract}
Hysteresis phenomena have been observed in different branches of physics and engineering sciences. Therefore several models have been proposed for hysteresis simulation in different fields; however, almost neither of them can be utilized universally. In this paper by inspiring of Preisach Neural Network which was inspired from Preisach model that basically stemmed from Madelung’s rules and using the learning capability of the neural networks, an adaptive universal model for hysteresis is introduced and called Extended Preisach Neural Network Model (XPNN). It is comprised of input, output and, two hidden layers. The input and output layers contain linear neurons while the first hidden layer incorporates neurons called Deteriorating Stop (DS) neurons, which their activation function follows DS operator. DS operator can generate non-congruent hysteresis loops. The second hidden layer includes Sigmoidal neurons. Adding the second hidden layer, helps neural network learn non-Masing and asymmetric hysteresis loops very smoothly. At the input layer, Besides, $x(t)$ which is input data, $\dot{x}(t)$, the rate at which $x(t)$ changes, is included as well in order to give XPNN the capability of learning rate-dependent hysteresis loops. Hence, the proposed approach has capability of the simulation of the both rate independent and rate dependent hysteresis with either congruent or non-congruent loops as well as symmetric and asymmetric loops. A new hybridized algorithm has been adopted for training of the XPNN which is based on combination of GA and the optimization method of sub-gradient with space dilatation. The generality of the proposed model has been evaluated by applying it on various hystereses from different areas of engineering with different characteristics. The results show that the model is successful in the identification of the considered hystereses. The proposed neural network shows excellent agreement with experimental data.
\end{abstract}


\section{Introduction} \label{sec:Introduction}

Hysteresis is a nonlinear phenomenon that appears in various systems including ferromagnetic materials, mechanical actuators and electronic relay circuits. Smart-materials-based actuators, such as piezoceramic actuators, magnetostrictive actuators and shape memory alloys, invariably show hysteresis effects. Hysteresis word first appeared in the literature in 1885 when Ewing used this Greek word in his paper on magnetism \cite{Ewing85}. The etymological meaning of hysteresis is “lagging behind.” It is known as a key component of hysteresis. Lagging, in a hysteretic system, can occur by changing the internal state and rate dependence. The latter means that the output of a rate-dependent system is related to the value of the input and the speed at which it changes. The rate dependence can cause the dynamic lag between input and output of the system. Often this kind of lagging is referred to as rate-dependent hysteresis. However, hysteresis usually in the literature means lagging behind not because of rate dependence so that Ewing introduced rate independence as another key aspect of hysteresis. Also, Visintin, \cite{visintin2013differential}, defined hysteresis as rate-independent memory effect. This definition disregarded any viscous-type effects which are coupled with typical hysteresis phenomena, such as ferromagnetism, ferro-electricity, and plasticity. However, in several cases rate independence assumption can be considered, provided that the evolution is not too fast. Hysteresis occurs in ferromagnetic, ferroelectric materials and magnetostrictive acuators, as well as in deformation of some materials such as metals in response to varying force. Many artificial systems are designed to have hysteresis. For example, in thermostats and Schmitt triggers, hysteresis is produced by positive feedback to avoid unwanted rapid switching. Hysteresis has been identified in many other fields, including economics and biology. Therefore, several researchers have proposed the rules and the models to simulate the different hysteretic phenomena. Among them, the researches done by Madelung, Masing, Prandtl and Preisach have been bases for the other researches in the literature.

Conventionally mathematical models with some free parameters are assumed for the simulation of the hysteresis material behaviors and their free parameters are tuned through identification. Unfortunately there is not a unique mathematical form that could be utilized for different materials since different materials do not have identical specifications. This is the main drawback of the mathematical models. Owing to this, Multilayer Feed-Forward Neural Networks (MFFNNs) as general approximator tools \cite{hornik1991approximation} have been found a special place in the identification of the hysteresis material behavior as model-free systems \cite{joghataie2012designing,joghataie2009nonlinear,joghataie2010transforming}.

On the other hand, conventional neural networks are static systems and do not have any internal memories \cite{hornik1991approximation}. Even though some works have been done to fix this problem, mathematically speaking, they are not suitable for simulating hysteresis behaviors. In order to compensate for this drawback of the conventional neural networks in hysteresis learning, there have been some attempts in literature to devise some new types of neural networks for the hysteretic problems \cite{joghataie2009nonlinear,joghataie2010transforming, joghataie2012designing,farrokh2016adaptive,joghataie2012neural}.

Inspired from prandtl-ishlinskii model, Joghataie and Farrokh \cite{joghataie2008dynamic} have proposed a new neural network called Prandtl Neural Network (PNN) and used it successfully in dynamic analysis of nonlinear inelastic frames and trusses \cite{joghataie2008dynamic,joghataie2011matrix}. PNN enjoys stop neurons in its hidden layer. Each stop neuron has a complete memory for hysteresis and an adaptive parameter which is tuned during training of the PNN. These two characteristics of stop neuron enable PNN to learn hysteretic behaviors without any data of the past state or non-measurable variables in its input layer; therefore, it is not subjected to error accumulation. However the PNN is suitable for the hysteretic problems without any degradation. Due to having some special properties, hysteresis loops of some materials are unusual which for instance are asymmetric, non-congruent and rate dependence under the cyclic and transient loading. As a result PNNs cannot learn them because they use stop activation functions which do not have capability of learning these special properties of those materials. Prandtl model only follows masing’s rules, so that the PNN cannot learn non-masing hysteresis behaviors and is merely suitable for the masing-congruent-rate independent hysteretic problems. Therefore, PNNs have not capability of learning non-congruent, non-masing and rate-dependent hysteresis behaviors. Stop neurons in the hidden layer of PNN only can create symmetric hysteresis loops which means they cannot learn asymmetric loops which is, for example, one of the characteristics of smart material hysteresis. Hence, PNNs are not suitable for modeling of material properties whose hysteresis behaviors are non-congruent, rate-dependent and asymmetric such as deteriorating hysteresis behavior of concrete and smart materials.

In 2015, authors \cite{farrokh2015modeling} have introduced a new activation function known as deteriorating stop (DS) with a non-congruent (or deteriorated) parameter, $\beta$, and used it instead of the stop activation function in the PNN. By this modification, the neural network is more general and it is called Generalized PNN (GPNN). The structure of GPNN is the same as Prandtl Neural Network (PNN) which was previously discussed but GPNN uses new neurons with activation function in accordance with DS operator. This newly defined neuron is called DS neuron. It enjoys non-congruency or deterioration behavior in its hysteresis loops; therefore, GPNN is able to model hysteresis behavior of materials whose their hysteresis loops are non-congruent. Although GPNNs are more generalized than PNNs and for instance are capable of learning non-congruent hysteresis, due to the properties of DS neurons they cannot learn non-masing behaviors such as asymmetric hysteresis loops and also rate-dependent hysteresis behaviors.  

Preisach model, which has been widely used for the modeling of hysteresis in different fields of engineering such as electromagnetism \cite{visone1998magnetic}, soil mechanics \cite{flynn2004application,guyer1995hysteresis} and shape memory alloys \cite{ortin1992preisach,zakerzadeh2010hysteresis}, has been developed based on Madelung's rules and incorporates relay operators. Preisach model is a rate independent hysteresis model which determines the output signal, y(t), where t represents time, through linear superposition of a set of continuous relay operators applied to the input signal, x(t). Russian mathematician Krasnoselskii separated this model from its physical meaning and stated it in pure mathematical form [12]. Afterwards, Mayergoyz \cite{mayergoyz1986mathematical}, based on phenomenological nature and mathematical generality of the Preisach model proved a theorem which gives the necessary and sufficient conditions for representation of actual hysteresis nonlinearities by the Preisach model based on deletion and congruency properties. The latter property means that all hysteresis loops corresponding to the same extremum values of the input are geometrically congruent. It can be concluded that the hysteretic behaviors which do not enjoy congruency property cannot be simulated by the Preisach model. 

Recently Farrokh and Joghataie have proposed a new type of multi-layer feed forward neural network (MLFFNN) by inspiring of Preisach model and it has been called Preisach Neural Network (Preisach-NN) \cite{farrokh2013adaptive}. It is comprised of input, output and two hidden layers. The input and output layers contain linear neurons while the first hidden layer incorporates neurons called stop neurons, which their activation function follows stop operator. The second hidden layer includes sigmoidal neurons. The optimization method of sub-gradient with space dilatation has been adopted for training of the Preisach-NN as a non-smooth problem. Preisach-NN not only has capability of learning hysteresis behaviors that follows masing’s rules, but also it has capability of learning non-masing hysteresis as well. Even though, the proposed Preisach-NN could be applied to most hysteresis behaviors, due to properties of the preisach model which are congruent and rate-independent, it cannot be used for hysteresis which are non-congruent and rate-dependent. 

In 2016, Farrokh and Dizaji \cite{farrokh2016adaptive}, proposed a new type of multi-layer feed forward neural networks by inspiring of madelung rules and it has been called neuro-Madelung model (NMM). Madelung's rules are common among most of the hysteresis phenomena \cite{brokate2012hysteresis}. The NMM has capability of simulating both rate-dependent and rate-independent hystereses with either congruent or non-congruent loops.  Using the learning capability of the neural networks, an adaptive general model for hysteresis is introduced according to the proposed approach. However, due to the nature of the neural network used in the NMM, and its input dependence on the input-output pair at the previous turning point and the input rate, it will be difficult to explicitly construct the inverse of the NMM \cite{farrokh2016adaptive}.

There are materials and actuators that exhibit rate dependent hysteresis, depending strongly upon the rate of change of the input in a highly nonlinear manner such as piezoceramic actuators, piezoelectric actuators and magnetostrictive actuators. In such cases, the Preisach and Prandtl–Ishlinskii models could yield considerable errors under inputs that are applied at varying rates. A few studies have proposed alternate density functions for enhancing the prediction abilities of these models under varying rates of input and output. Mayergoyz\cite{mayergoyz1986mathematical,mayergoyz2003mathematical} proposed a rate dependent Preisach model by introducing the speed of the output in the density function. Tan and Baras \cite{tan2004modeling,tan2005adaptive} integrated the dynamic equilibrium equation of a single degree of freedom with the Preisach operator to characterize and to compensate the rate-dependent hysteresis in a magnetostrictive actuator at or above 10 Hz.

Smith \cite{smith2003free,smith2006unified,smith2012homogenized} presented a homogenized energy model using the Preisach model to characterize the rate-dependent hysteresis up to 1–2 kHz excitation frequency. An alternate rate-dependent Preisach model comprising a modified relay operator was also proposed by Bertotti \cite{brokate2012hysteresis} to model the rate-dependent hysteresis of soft magnetic materials at different frequencies. The operator function was formulated to relate the rate of change of the flux to differences between the input and the Preisach operator thresholds. The model results showed reasonably good agreement with the experimental data and that the area within the hysteresis loops increased with an increase in the frequency of the input magnetic field. The reported modifications are generally focused on the Preisach model by including the speed of the input in the density functions, which may only offer limited ability to describe the rate-dependent hysteresis. 

As it can be seen from the literature, each of the previously proposed models have their own limitations. For brevity, PNNs only follows masing’s rules and are rate-independent. Therefore they have not capability of simulating non-congruent, rate-dependent and asymmetric hysteresis. Even though GPNN has solved the issue of non-congruency and can simulate hysteresis whose behaviors are non-congruent, it has limitation of learning non-masing and rate-dependent hysteresis. Preisach-NN has capability of learning non-masing hysteresis. On the contrary, it is limited to learn congruent hysteresis which means that it has not capability of learning non-congruent hysteresis behaviors. 

In this paper, a new type of multi-layer feed forward neural network has been proposed by inspiring of Preisach-NN and is called Extended Preisach Neural Network (XPNN). The proposed model, is more general than Preisach-NN in terms of the congruency and rate dependency standpoint. The generality of the proposed model has been evaluated by applying it on various hystereses from different areas of engineering with different characteristics. The results show that the model is successful in the identification of the considered hystereses. Also, the model shows excellent agreement with experimental data.

The structure of the paper is as follows: Firstly hysteresis phenomena and its several properties is investigated and after a short review on the PNN, GPNN and Preisach-NN, the new proposed neural network is introduced with more precise detail. Its architecture and training algorithm have been discussed. The capability of the new proposed training algorithm is shown. Finally the application of the XPNN in learning is explained using different hysteresis behaviors.
\section{Hysteresis rules and models} \label{sec:Hysteresis}

Hysteresis phenomenon is seen in many branches of science and engineering such as chemistry, biology, mechanics, electronics, and optics \cite{Ewing85,morris2011hysteresis}. Hence different rules and models for hysteresis have been reported in the literature. Amongst them, the Madelung's rules, Masing, Prandtl, and  Preisach models \cite{mayergoyz2003mathematical} are very popular and related to this paper issue. In the following the description of these rules and models is introduced.

\subsection{Madelung's rules}
\begin{figure}[h!]
	\centering
	\includegraphics{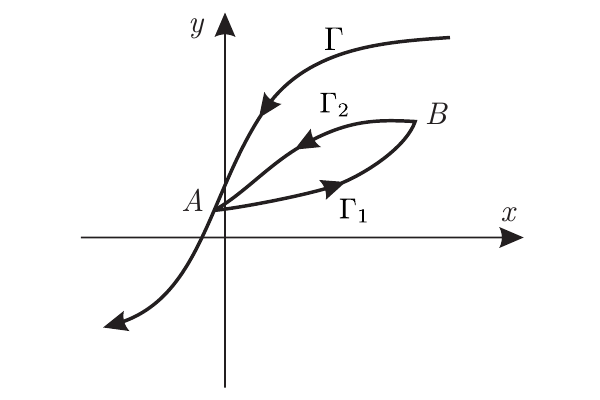}
	\caption{Madelung's rules.}\label{Madelung-fig}
\end{figure}

In the 1900s, Madelung tried to formalize the rules governing the experimentally observed branchings and loopings of ferromagnetic hysteresis in his dissertation \cite{brokate2012hysteresis}. Consequently he proposed the three following rules:
1) Any curve $\Gamma_1$ emanating from a turning point A in Fig.~\ref{Madelung-fig} is uniquely determined by the coordinates of the A.
2) If any point B on the curve $\Gamma_1$ becomes a new turning point, then the curve originating as B leads back to point A. 
3) If the curve $\Gamma_2$ is continued beyond the point A, then its coincides with the continuation of the curve Γ which led to the point A before the $\Gamma_1-\Gamma_2$-cycle was traversed.

Actually Madelung's rules describe the deletion or wiping out property of the hysteresis memory. According to these rules, nested hysteresis loops can be generated in the input-output plane. This property is not just belong to the field of ferronagnetism and the same behavior in the hysteresis of the other fields has been observed so that the deletion known as the most common property between different hysteretic phenomena.

\subsection{Masing model}
Masing \cite{masing1926eigenspannumyen} in a research about brass developed some rules for hysteresis of metal under cyclic loading. Masing asserted that if the load-deflection curve for entire system at the virgin loading is odd symmetric and is given by:
\begin{equation}
y=g(x) \label{masing1}
\end{equation}
where $x$ is displacement (input) and y is restoring force (output), then unloading and reloading branches of hysteresis loops for steady-state response are geometrically similar to the virgin loading curve and are described by the same equation except for two-fold magnification:
\begin{equation}
y=y^*+2g(\dfrac{x-x^*}{2})\label{masing2}
\end{equation}
where $(x^*,y^*)$ is the last load reversal point. As the Masing's rules originally was developed for steady-state (cyclic) loading, it cannot be utilized for transient loading in which the minor loops in the hysteresis might be generated and closed by backing to the emanation point of the minor loop and continuing the path that would be traversed in the absent of the minor loop \cite{masing1926eigenspannumyen}. There have been some attempts in the literature to extend the rules for transient cases \cite{jayakumar1987modeling}. Actually this extension is a combination of the Madelung's rules and original Masing's rules. Based on the extended Masing's rules, several hysteresis models have been proposed and almost all of them can simulate the symmetric hysteresis behaviors.

\subsection{Prandtl model}
The Prandtl model is a superposition of elementary stop operators, which are parameterized by a single threshold variable. This model is defined in terms of an integral of stop operator with a density function determining the shape of the hysteresis loop as: 
\begin{equation}
y(t)=\int_{0}^{\infty} w(r) \varepsilon_r\left[x(t)\right] dr \label{eq-prandtl}
\end{equation}
where $w(r)$= density function;  $\varepsilon_r[.]$= stop operator; $r$= stop operator thershold; $x(t)$= input signal; and $y(t)$= output signal. The discrete from of the Prandtl model known as Prandtl--Ishilinskii and is:

\begin{equation}
y(t)=\sum _{j=1}^{m}{w}_{j}{\varepsilon }_{{r}_{j}}[x(t)]\label{eq-ishlinskii}
\end{equation}

\begin{figure}
	\centering 
	\includegraphics{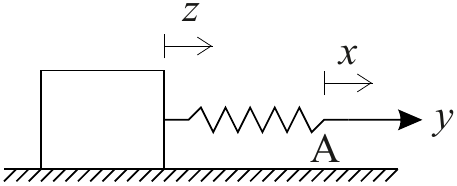} 
	\caption{Frictional sliding of a body-spring system along a surface describing stop and play operators.}
	\label{stop-mechanism-fig} 
\end{figure}

The stop operator, $y_r(t)=\varepsilon_r[x(t)]$, can be physically interpreted as the mechanism shown in Fig.~\ref{stop-mechanism-fig}, in which $x(t)$= displacement of the spring end A; $r$= Coulomb friction force between the body and the surface; and $y_r(t)$ is internal force of the spring or equivalently
relative deformation of the spring ends (because the stiffness of the spring is assumed to the value of 1). The mathematical definition
of the stop operator on each subinterval $[t_{i},t_{i+1}]$ where
$x(t)$ is piecewise monotone on $0=t_{0}<t_{1}<t_{2}<\ldots<t_{N}=t_{E}$,
$y_{r}(t)=\varepsilon_{r}[x](t)$ can be described by induction as \cite{brokate2012hysteresis}:
\begin{equation}
\begin{gathered}y_{r}(0)=e_{r}(x(0))\\
y_{r}(t)=e_{r}(x(t)-x(t_{i})+y_{r}(t_{i}))\\
\quad\mbox{for}\quad t_{i}<t\le t_{i+1};\:0\le i\le N-1\end{gathered}
\end{equation}
where \begin{equation}
e_{r}(v)=\min\{r,\max\{-r,v\}\}.
\end{equation}

As each stop operator in Eqs.~(\ref{eq-prandtl}) and (\ref{eq-ishlinskii}) is rate-independent, the Prandtl model generates the rate independent hysteresis. It means that the rate of the input signal does not affect the output signal. Also each stop operator in the Prandtl model complies both Madelung's and Masing's rules; therefore, the Prandtl model inherits these properties from the stop operators by linear combination. Hence the Prandtl model could be successfully utilized for the hystereses enjoy the extend Masing's rules. 

\subsection{Preisach model}
The Preisach hysteresis model is perhaps the most well-known phenomenological operator-based model for characterizing hysteresis phenomenon. In the Preisach formulation, hysteresis is modeled as a cumulative effect (density function) of all possible delayed relay elements (relay operators), which are parameterized by a pair of threshold variables \cite{brokate2012hysteresis}. This model is completely characterized by two properties: wiping out and congruent minor loop properties.

\begin{figure*}[!t]
	\centering
	\includegraphics[width=5in]{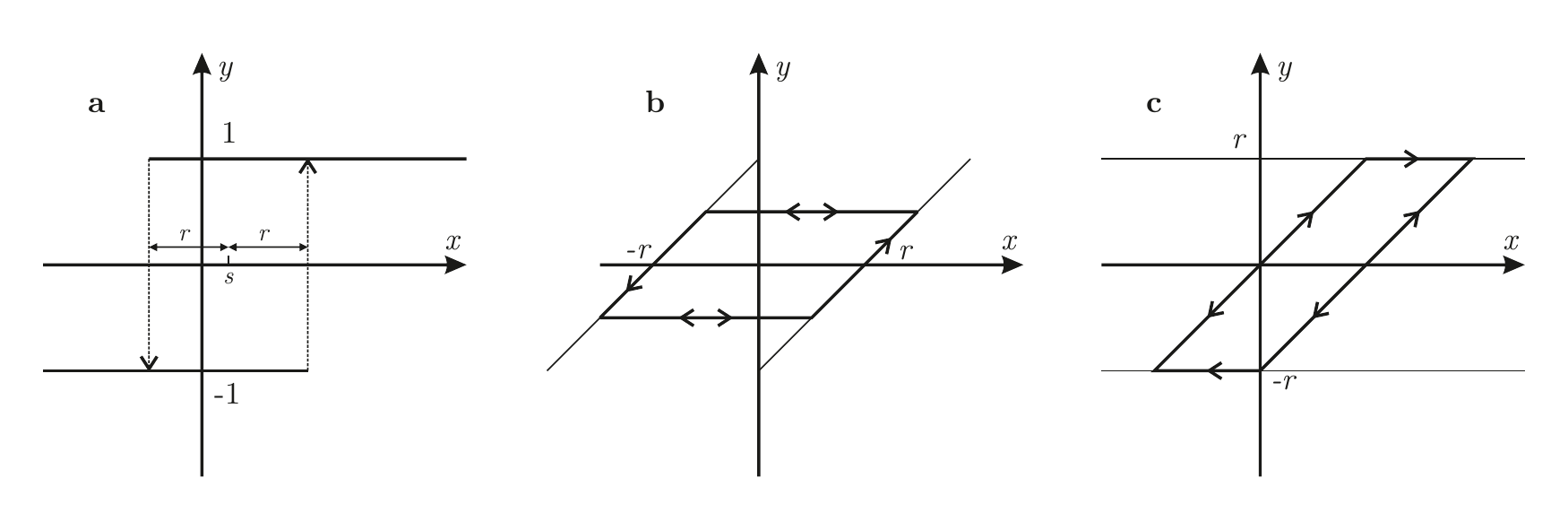}
	\caption{Hysteresis operators: a)Relay operator b) Play operator c) Stop operator}
	\label{fig-operator}
\end{figure*}

Preisach model has been widely used for the modeling of hysteresis in different fields of engineering such as electromagnetism \cite{visone1998magnetic}, soil mechanics \cite{guyer1995hysteresis}, and shape memory alloys \cite{ortin1992preisach,zakerzadeh2010hysteresis}. It has been developed based on Madelung’s rules \cite{brokate2012hysteresis} and incorporates relay operators. Preisach model is a rate-independent hysteresis model which determines the output signal $y(t)$ where $t$ represents time, through linear superposition of a set of continuous relay operators applied to the input signal $x(t)$ as follows:

\begin{equation}
y(t)=\int _{0}^{\infty}\int _{-\infty}^{\infty}\mu(r,s){R}_{s-r,s+r}[x](t)dsdr \label{eq-preisach}
\end{equation}

where $\mu(r,s)$ density function and ${R}_{s-r,s+r}=$ relay operator with mean value of $s$ and half-width value of $r$ shown in Fig.~\ref{fig-operator}a. The initial values of the relays are taken as $-1$, if $s>0.0$, and as $+1$, otherwise. There is a curve, $\psi(t,r)$, which divides the half-plane $r>0$ into two distinct regions so that the value of all the relays is $+1$ in the lower region and $-1$ in the upper region. Mathematically it can be shown that $\psi(t,r)$ is a \emph{play} operator as follows \cite{brokate2012hysteresis}:
\begin{equation}
\psi(t,r)=P_{r}[x](t)  \quad \forall r\ge 0
\end{equation}
where $P_r[x](t)$ has been depicted in Fig.~\ref{fig-operator}b. Hence according the former explanations the Preisach model can be written as:
\begin{equation}
y(t)=\int_{0}^{+\infty}\left[\int_{-\infty}^{\psi(t,r)}\mu(r,s)\mathrm{d}s-\int_{\psi(t,r)}^{+\infty}\mu(r,s)\mathrm{d}s\right]\mathrm{d}r
\end{equation}
Eventually defining,
\begin{equation}
f(r,u)=\int_{-\infty}^{u}\mu(r,s)\mathrm{d}s-\int_{u}^{+\infty}\mu(r,s)\mathrm{d}s
\end{equation}
Eq.~(\ref{eq-preisach}) can be rewritten in its general form as follows:
\begin{equation}
y(t)=\int_{0}^{\infty}f(r,\psi(t,r))\mathrm{d}r
\label{eq-preisach1}
\end{equation}
According to Eq.~(\ref{eq-preisach1}) the model has a memory which consists of a continuous set of play operators. Therefore the Preisach model can be decomposed into a memoryless functional and a continuous set of play operators. In general the input-output relation of the model is as follows:
\begin{flalign}
y(t) &=Q\left[\psi(t,r)\right] \label{eq-memoryless}\\ 
\psi(t,r) &=P_r[x](t) \quad \forall r \ge 0 \label{eq-memory}
\end{flalign}
where $Q[\psi(t,r)]$ is a memoryless functional which maps the function $\psi(t,r)$ into the real valued output $y(t)$. \emph{Stop} operator is another operator (Fig.~\ref{fig-operator}c) usually used in hysteresis simulation contexts. As stop and play operators can be converted to each other through 
\begin{equation}
P_r[x](t)=x(t)-\varepsilon_r[x](t) \label{eq-play-stop}
\end{equation}
where $\varepsilon_r[.]$ denotes a stop operator with threshold $r$, the memory of the Preisach model also can be defined in terms of stop, instead of play operators. Comparison of Eqs.~(\ref{eq-preisach1}) and (\ref{eq-prandtl}) indicates that the Prandtl model is a special case of the Preisach model because Eq.~(\ref{eq-preisach1}) is mathematically more general than Eq.~(\ref{eq-prandtl}). Owing to this the Preisach model can be utilized in the simulation of the hysteretic behavior which does not follow Masing’s rules. 

Mayergoyz (1986) \cite{mayergoyz1986mathematical} showed that all hysteresis loops generated by Preisach model with the same extermum values of input $x$ are geometrically congruent. In addition, the model memory enjoys the deletion property described by Madelung. Congruency means that the loops with the identical extermum have the same shape. The Preisach model ensures this properties. Therefore, Preisach model cannot applied to hysteresis systems which their loops and also their sub-loops are non-congruent.

\section{New neural networks to simulate hysteresis behaviors} \label{sec:NeuralNetwork}
Artificial neural networks as learning systems have been found special place  in hysteresis simulation and compensation. However, these conventional neural networks such as recurrent and time delay neural networks do not have complete capability in learning of a hysteretic behavior \cite{wei2000constructing}. In order to compensate for this drawback of the conventional neural networks in hysteresis learning, there have been some attempts in literature to devise some new types of neural networks for the hysteretic problems. Recently Joghataie and Farrokh have proposed a new neural network called Prandtl neural network (PNN) and used it successfully in hysteresis behaviors which obey from Masing rules. In the another attempt by the Farrokh et al. (2015) \cite{farrokh2015modeling} a new type of NN is introduced by which non-congruent hysteresis behaviors could be simulated and they called it GPNN. In the last attempt to create a NN to simulate non-Masing hysteresis behaviors, Farrokh and Joghataie proposed a Preisach-NN which could simulate non-Masing hysteresis behaviors precisely. However most of the smart materials hysteresis have different characteristics such as  rate-dependency, asymmetric and in general do not obey from aforementioned defined rules for hysteresis such as Masing rules hysteresis. Each of the proposed neural networks have their own limitation in learning hysteresis behaviors and none of them are universal in having capability of learning all hysteresis behaviors. On the other hand, our purpose is introducing a universal neural network which would be able to learn all different properties of the hysteresis simultaneously. 

\subsection{Prandtl neural networks (PNNs)}
\begin{figure}[ht!]
	\centering
	\includegraphics[width=3in]{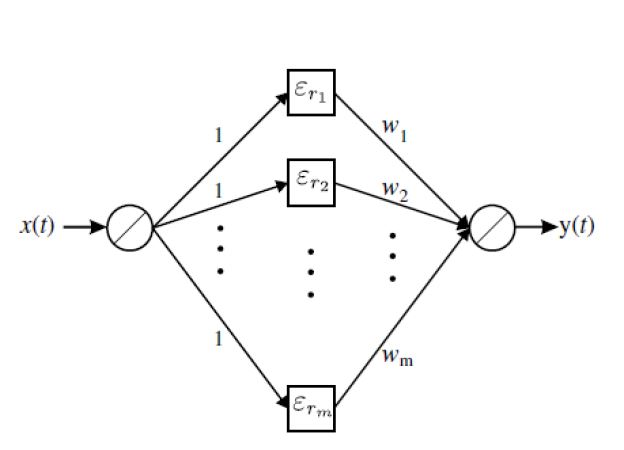}
	\caption{Prandtl Neural Networks (PNNs) architecture}
	\label{PNN-Arc}
\end{figure}

Prandtl neural network is firstly introduced by Joghataie and Farrokh in 2008 \cite{joghataie2008dynamic}. It has been inspired by Prandtl-Ishlinskii model.  The mathematics of the Prandtl neural network is as Eq.~(\ref{eq-ishlinskii}). The equation could be shown as a multi-layer feed-forward neural network according to Fig.~\ref{PNN-Arc} that includes input and output layers with linear neurons and a hidden layer with several stop neurons \cite{joghataie2008dynamic}. The activation function of each stop neuron is identical to the stop operator. There are $2m$ free parameters in a Prandtl-NN with $m$ stops neurons in its hidden layer which are tuned during its training.  Two types of Prandtl-NN have been proposed by Joghataie and Farrokh. In type 1 \cite{joghataie2011matrix}, the connection weights between input and hidden layers are assigned to the value of one and are not tuned during the training. However, the threshold values of stop neurons in hidden layer and the connection weights between hidden and output layers are determined by training. On the contrary, in type 2 \cite{joghataie2011matrix} the normalized stop neurons have the unit threshold value while the input-hidden connections are determined with training algorithm. This type is more like conventional feed-forward neural networks and the input layer neurons can be increased easily if it is needed.  The training algorithm for Prandtl-NNs is composed of genetic algorithm and least squares \cite{joghataie2008dynamic, joghataie2011matrix}. As the mathematics of the Prandtl-NNs are the same as the Prandtl-Ishlinskii model, they include all hysteresis properties of the Prandtl-Ishlinskii model. Consequently it is only capable to learn symmetric, congruent, and rate-independent hysteresis. In addition, they can be utilized only for hystereses that obey the Masing's rules \cite{farrokh2013adaptive}.  

\subsection{Generalized prandtl neural network (GPNNs)}
In 2015, Farrokh and his co workers \cite{farrokh2015modeling} generalized Prandtl-NN for the simulation of the deteriorating hysteresis behaviors.The structure of generalized Prandtl-NN is the same as Prandtl-NN type 1 which was previously discussed except that it uses new neurons in its hidden layer. This newly defined neuron has been called Deteriorating Stop (DS) neuron. It can generate the hysteresis loops with both stiffness and strength degradation.  The physical mechanism of a DS neuron is the same as the stop operator shown in Fig.~\ref{stop-mechanism-fig}. However, the slip of the body along the surface in Fig.~(\ref{stop-mechanism-fig}) causes some damages in the mechanism of the DS neuron contrary to the original stop neuron. The cumulative
slip of the body can be calculated in terms of the play operator by the following induction \cite{farrokh2015modeling}: 
\begin{equation}
s_{r}(t_{i+1})=|{P_{r}[x(t_{i+1})]-P_{r}[x(t_{i})}]|+s_{r}(t_{i})\label{eq-slip}
\end{equation}
where $s_{r}(t_{i})$ is the cumulative slip at $t_{i}$ with $s_r(t_{0})=0$.
By defining a soundness index in terms of the cumulative slip the output
of Deteriorating Stop (DS) operator is obtained. Here the simplest
function for the soundness index (i.e. bilinear form) has been considered for DS operator by the
following equation, 
\begin{equation}  
BL_{r,\beta}(s_{r})=\begin{cases}
1-\dfrac{s_{r}}{r\beta} & s_{r}<r\beta\\
0 & s_{r}>r\beta\end{cases}\label{eq-bl}
\end{equation}
where $\beta$ is a parameter that control the rate of the deterioration
in the DS operator so that the small value for $\beta$ expedites
the deterioration process in DS operator. Therefore the mathematical
form of the DS operator, $DS_{r,\beta}$, can be defined by:

\begin{equation}
DS_{r,\beta}[x(t)]=\varepsilon_{r}[x(t)]\times BL_{r,\beta}(s_{r})\label{eq-ds}
\end{equation}
Hence the generalized Prandtl-NN has been mathematically stated by Eq.~(\ref{eq-gpnn}):

\begin{equation}
y(t)=\sum _{j=1}^{m}{w}_{j}{DS}_{{r}_{j},{\beta}_{j}}[x(t)]\label{eq-gpnn}
\end{equation}
Eq.~(\ref{eq-gpnn}) could be represented in the form of a multilayer feed-forward neural network with linear neurons in the input and output layers and $m$ DS neurons in the hidden layer. The connection weights between the input neuron and all neurons in the hidden layer have been assumed to have a value of 1.0 because of the equivalence with Eq.~(\ref{eq-gpnn}). In a generalized Prandtl-NN with $m$ DS neurons in the hidden layer, there are totally $3m$ free parameters of which $m$ parameters are the connection weights between the hidden layer neurons and the output neuron ($w_j$; j = 1, 2, ... , $m$) and $2m$ parameters are the internal free parameters for DS neurons in the hidden layer ($r_j$, $\beta_j$; $j$ = 1, 2,..., $m$). It should be mentioned that $\beta_j$'~s can control overall non-congruency deterioration of the generalized Prandtl-NN so that if all the $\beta_j$~'s have the large values the deteriorating capability will diminish and hence it will behave like a Prandtl-NN. Genetic algorithm and least squares have been used for tuning DS neurons parameters ($r_j$ and $\beta_j$) and hidden-output connection weights, respectively.  

\subsection{Preisach neural network}
\begin{figure} 
	\centering
	\includegraphics[width=3in]{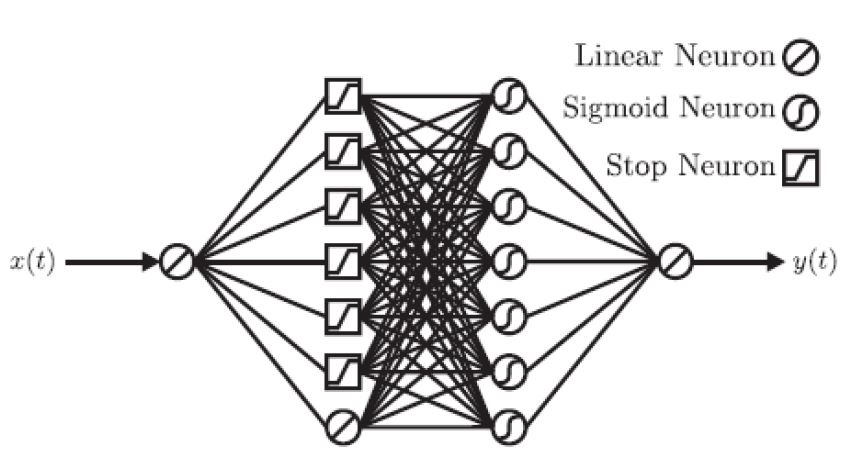}
	\caption{Preisach Neural Network architecture}
	\label{Preisach-NN}
\end{figure}

In 2014, Farrokh and Joghataie \cite{farrokh2013adaptive} proposed a new type of multilayer feed-forward neural network by inspiring of Preisach model and it has been called Preisach Neural Network (Preisach-NN). It is comprised of input, output and two hidden layers. The input and output layers contain linear neurons while the first hidden layer incorporates normalized stop neurons, which their activation function follows stop operator with threshold  value of one. The second hidden layer includes sigmoid neurons. Equations (\ref{eq-memoryless}) and (\ref{eq-memory}) are the concept of the Preisach-NN so that the first and second hidden layer play the role of Eqs. (\ref{eq-memory}) and (\ref{eq-memoryless}), respectively. Details of Preisach-NN can be seen in Fig.~\ref{Preisach-NN}.  Its training algorithm is based on the method of the sub-gradient with space dilatation. Although the proposed Preisach-NN could be mathematically identical to Preisach model, tuning the Preisach-NN is easier and also more general than the model. Some Preisach-NNs are trained successfully to learn two different types of synthetically generated Masing and non-Masing hysteresis problems with very high precision. The Preisach-NN the and as its origin has capability of learning Masing hysteresis as well as non-Masing hysteresis behaviors with either symmetric or asymmetric hysteretic loops. Although Preisach-NN can be applied for most family of hysteresis, but they cannot learn non-congruent as well as rate-dependent hysteresis. It is obvious that if the second hidden layer of Preisach-NN is omitted, it is converted to Prandtl-NN type 2. It shows that Preisach-NN is more general than Prandtl-NNs.

\section{Extended preisach neural networks (EPNNs)} \label{sec:Extended}

\subsection{Architecture}
\begin{figure}[h!]
	\centering
	\includegraphics[width=3in]{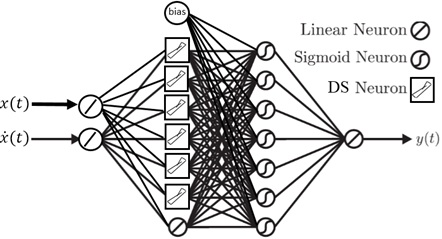}
	\caption{Extended Preisach Neural Network architecture}\label{EPNN}
\end{figure}

The previously proposed neural networks; Prandtl-NNs, generalized Pradntl-NN, and Preisach-NN have their limitations and their usage need some knowledge about physical properties of considered hysteresis because neither of them is  generally able to simulate hysteresis phenomena. Moreover, none of them have the capability in simulation rate-dependent hysteresis. In this paper a new neural network is introduced by inspiring the aforementioned neural networks and it is enhanced for the rate-dependent cases. This new kind of neural network called extended Preisach neural network because its architecture basically is the same as Preisach-NN which is emanated from Preisach model. Similar to Preisach-NN, it is comprised of input, output and two hidden layers. The input and output layers contain linear neurons while unlike Preisach-NN, the first hidden layer incorporates Normalized Deteriorating Stop (NDS) neurons, which their activation function follows deteriorating stop operator with unit threshold ($r=1$). Moreover, in the input layer of the extended Preisach-NN in addition to the linear neuron for input signal $x(t)$, an extra neuron for the rate of the input signal $\dot{x}(t)$ has been considered. Details of extended Preisach-NN architecture has been shown in Fig.~\ref{EPNN}. It is obvious that the architecture of the extended Preisach-NN has been considered so that it encompasses all specifications of the previously proposed Prandtl-NNs, and Preisach-NN in unique structure. Furthermore, it is a rate-dependent neural network while neither of the aforementioned neural networks could be utilized for the rate-dependent cases. 

\subsection{NDS neuron} 
\begin{figure}
    \centering
     \includegraphics[width=\linewidth]{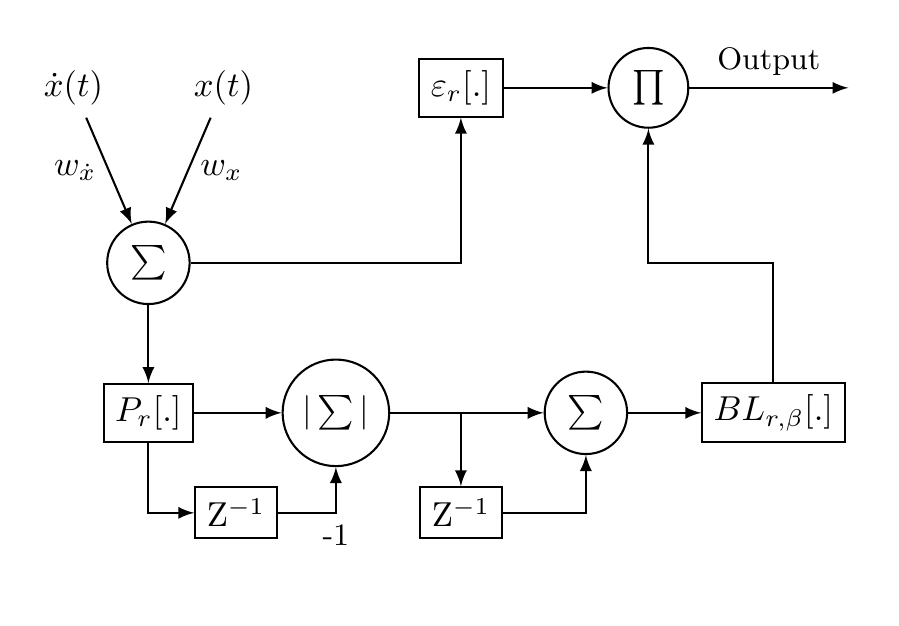}
    \caption{Block diagram of the internal computation of a NDS neuron ($r=1$).\label{NDS-fig}}
\end{figure}

NDS neurons are the most important units of the extended Preisach-NN and they make the proposed neural network be different from the conventional MLFFNNs. Hence understanding of their properties could be instructive. Figure~\ref{NDS-fig} shows the internal computation of a NDS neuron in the form of the block diagram. Each NDS neuron according to the architecture shown in Fig.~\ref{EPNN} receives signal from the input layer neurons. In Fig.~\ref{NDS-fig}, $w_x$ and $w_{\dot{x}}$ are the connection weights for $x(t)$ and $\dot{x}(t)$ input neurons, respectively. 

\subsection{Static and Dynamic Loops} 
Lagging, in a hysteretic system, can occur by changing the internal state and rate dependence. The latter means that the output of a rate-dependent system is related to the value of the input and the speed at which it changes. The rate dependence can cause the dynamic lag between input and output of the system. Often this kind of lagging is referred to as rate-dependent hysteresis. The second kind of lagging behind is not because of rate dependence so that Ewing introduced rate independence as another key aspect of hysteresis. Therefore, in order to introduce a comprehensive model, we should include both type of lagging, in a hysteresis system, in the proposed model. Many models have been proposed, but they are not able to include both of properties in themselves. 

\begin{figure}
	\centering
	\includegraphics[width=\linewidth]{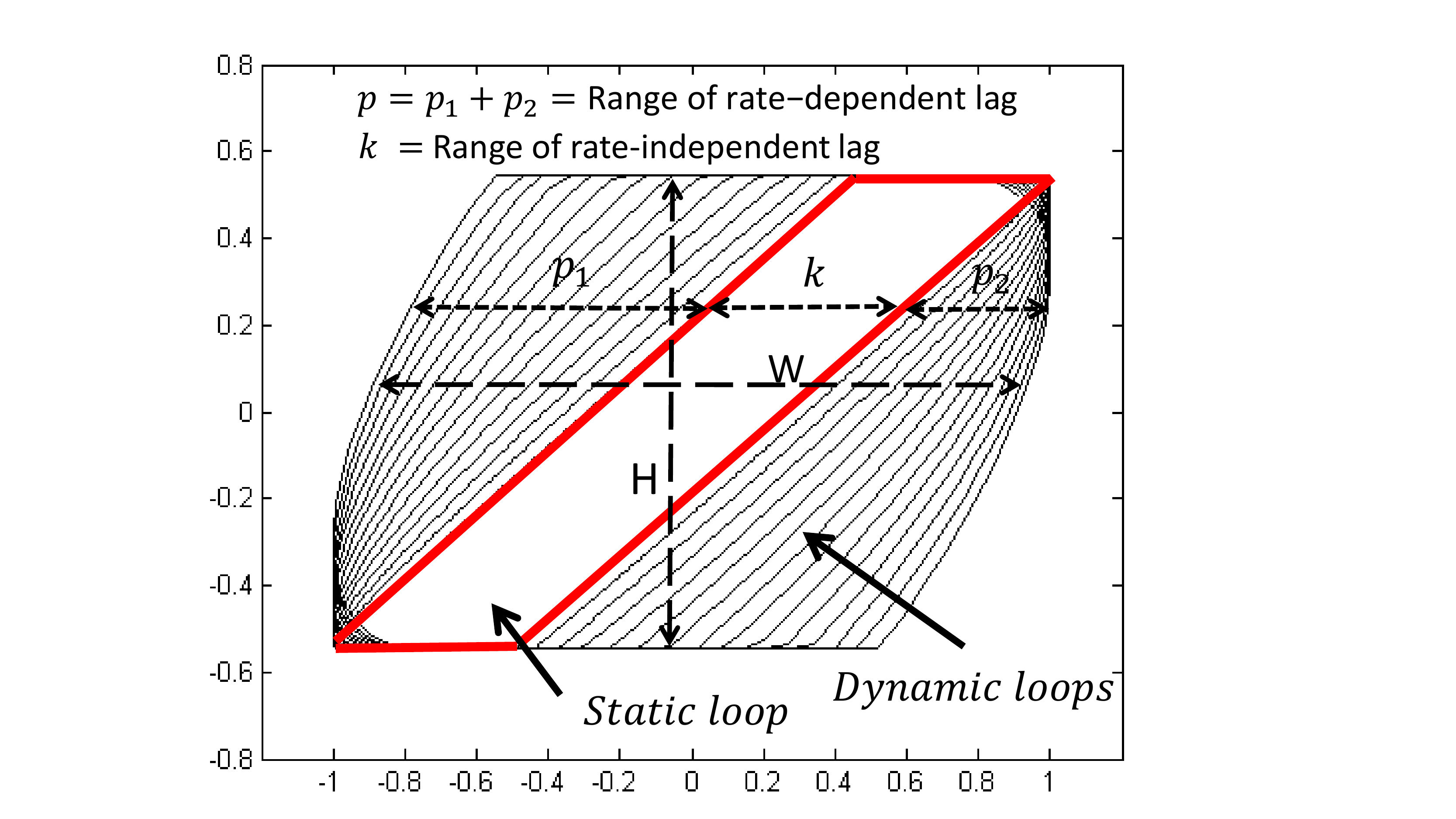}
	\caption{Dynamic and Static loops}\label{Dynamic}
\end{figure}

On the other hand, the proposed model in this paper have concluded both lagging in itself.  In order to show both lagging in the proposed model, in the Fig.~\ref{Dynamic} two type of lagging have been shown. As it can be seen, the hysteresis loop has two part. One part has to do with lagging behind generated by internal state which is rate-independent part and second part which has to do with lagging behind caused by speed changing of input which is called rate-dependency part.

As it can be seen from Fig.~\ref{Dynamic}, $W$ and $H$ can be changed as frequency of the inputs changes, which is properties of the most of the smart materials, specially the magnetostrictive actuators. 
Weight connections of $w_1$ and $w_2$ controls $H$ and $W$; with changing $w_1$ and $w_2$, $H$ and $W$ changes as well.

In the Fig.~\ref{Dynamic}, hysteresis loops have been separated into two cases: static hysteresis loop and dynamic hysteresis loops. Static loop is rate-independent and it only related to mechanical lags which can be generated from internal changing of the system. Dynamic loops are rate-dependent and contingent upon rate at which input data changes. Therefore, if the rate at which input of the system changes suddenly, the most part of the hysteresis loops would be dynamical. On the other hand if the rate of the input changes smoothly, there would be static loop and the system would be rate-independent. In the Fig.~\ref{Dynamic}, the range of rate-independency and the range of rate-dependency are k and p respectively.  The values of the k and p depend on the w1 and w2 respectively.

If $w_1$=0.0 or less than maximum pick of the input signal and $w_2$$\neq$0.0, then $k$=0.0 and $p$$\neq$0.0, therefore, the system would be rate-dependent.
If $w_2$=0.0 and $w_1$$\neq$0.0 which is higher than maximum of the input signal, then $p$=0.0 and $k$$\neq$0.0 therefore, the system would be rate-independent.
If $w_1$$\neq$0.0 and $w_2$$\neq$0.0, then $k$$\neq$0.0 and $p$$\neq$0.0, therefore, the system would be both rate-independent and rate-dependent. Therefore, it can be concluded that every system can be either rate-dependent or rate-independent or both. 

Defining such models which could have capability of distinguishing these characteristics would be a very cumbersome work.  The proposed model in this paper has capability of distinguishing the manner of system automatically. In other words, it is capable of realizing that whether our system is rate-independent, rate-dependent or both.  

Therefore it can be concluded that the proposed model is very suitable to simulate rate-dependent hysteresis behaviors.

\subsection{An Example-NDS neuron with a rate dependent signal} 
\begin{figure*}
	\centering
	\includegraphics[height=3in,width=6.0in]{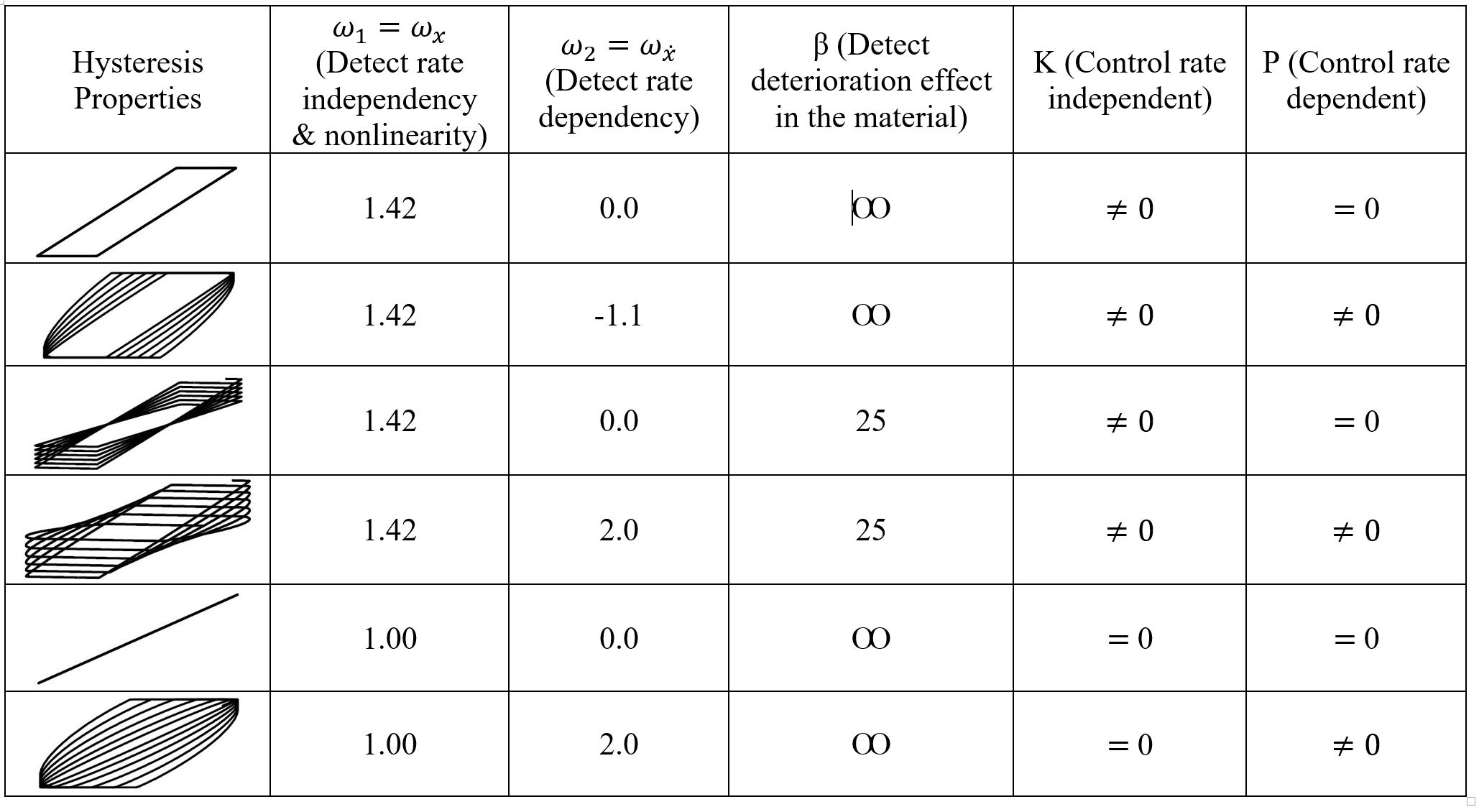}
	\caption{The summary of 6 different cases of hysteresis properties}\label{Capture}
\end{figure*}

\begin{figure}[t]
	\centering
    \includegraphics[height=5in,width=3in]{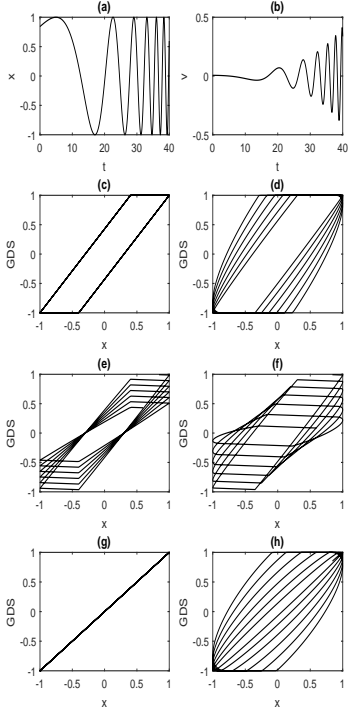}
	\caption{Different loops}\label{loops}
\end{figure}

In order to investigate the properties of a NDS neuron an example of input signal and its rate have been considered as Fig.~\ref{loops}(a-b). The sensitivity of the NDS neuron output to the values of $w_x$, $w_{\dot{x}}$ and $\beta$ is discussed in the following. For the input signal $-1 < x(t) < +1$ as shown in the Fig.~\ref{loops}a three cases are considered as follows: Case 1: if $-1 \leqslant w_1 \leqslant 1$ and $w_2$$\neq$ 0.0 then K = 0 and $w_2$$\neq$ 0.0; case 2: if $w_1$<-1 or $w_x$>1 and $w_2$ = 0 then $K$$\neq$ 0.0 and P = 0; case 3: if $w_1$<-1 or $w_x$>1 and $w_2$$\neq$ 0.0 then $K$$\neq$ 0.0 and $P$$\neq$ 0.0. The definition of K and P can be found from Fig.~\ref{Dynamic}. In the first case the values of $w_x$, $w_{\dot{x}}$ and $\beta$ are 1.42, 0.0, and $1\times 10^7$, respectively. The reason that I selected value higher than 1 for $w_x$ is that maximum value of the input signal is 1, so in order to show hysteresis behaviour, the value of $w_x$ should be defined higher than 1. In this case the hysteresis loops are rate-independent because $w_{\dot{x}}=0.0$. Hence it acts like stop operator. The hysteresis loops generated by the NDS neuron in response to the assumed input signal have been shown in Fig.~\ref{loops}c. The hysteresis loops are symmetric,  congruent, and rate-independent. In the second case the values are 1.42, -1.1, 10000000 respectively. In this case as you can see in the Fig.~\ref{loops}d, because of the factor $w_2$, the hysteresis loops have created different loops depend on how fast the rate of the x(t) changes. Therefore it can be concluded that installing \.{x}(t) at the input of the neuron, turn the neuron into rate-dependent operator. Also based on Fig.~\ref{loops}d, while the frequency of input increases, the hysteresis loops is going to get far from each other which means with increasing w2, rare-dependency property also enhances, which in turn is in compatible with most real rate-dependent hysteresis loops behaviors. In the third case, the values of $w_1$, $w_2$ and $\beta$ are 1.42, 0.0, 25 respectively.  At it can be seen in Fig.~\ref{loops}e, in this case, due to $w_2$ = 0.0, effect of rate-dependency has been eliminated; However due to $\beta$ = 50, the effect of non-congruency is obvious; therefore in this case hysteresis loops are rate-independent and non-congruent. In the forth case the values of $w_1$, $w_2$ and $\beta$ are 1.42, 2 and 25 respectively. This case is the most complete case in which the hysteresis loops not only have rate-dependency effect but also they are non-congruent (Fig.~\ref{loops}f). In the fifth case the values of $w_1$, $w_2$ and $\beta$ are 1, 0.0 and 10000000 respectively. In this case due to the values of $w_1$ and $w_2$ there in neither static hysteresis loop nor dynamic hysteresis loops. In another words because of $w_1$=1, there would be no static hysteresis loop which in turn shows the parameter $w_1$ is in charge of static loops (Fig.~\ref{loops}g). In the last case the values of $w_1$, $w_2$ and $\beta$ are 1, 2 and 10000000 respectively. In this case hysteresis loops only are dynamic or rate-dependent. In another words due to $w_2$=2, there would be only dynamic hysteresis loops (Fig.~\ref{loops}h). The summary of the 6 cases can be seen in Fig.~\ref{Capture}.

\subsection{Detection of system characters} 
Using the values for $w_1$, $w_2$ and $\beta$, we can detect properties of hysteresis of materials. For instance based on table 1 different properties of input signals can be detected.

\begin {table}[H]
\caption {different properties of input signals } \label{table1}
\begin{center}
	\begin{tabular}{|l||c|c|c|c|c|}
		\hline Hysteresis properties& $w_1$ & $w_2$ & $\beta$ & K & P \\ 
		\hline \hline Masing & 1.42 & 0 & Y & N & 0 \\ 
		\hline Non-masing & N & N & Y & Y & Y \\ 
		\hline Symmetric & Y & Y & Y & Y & Y\\ 
		\hline Asymmetric & N & N & Y & Y & Y\\ 
		\hline Rate-independent & Y & Y & Y & Y & Y\\ 
		\hline Rate-dependent & N & N & N & Y & Y\\ 
		\hline Congruent & Y & Y & Y & Y & Y\\ 
		\hline Non-congruent & N & Y & N & Y & Y\\ 
		\hline Algorithm & GA & GA & SGSD & HA & Y\\ 
		\hline 
	\end{tabular} 
\end{center}
\end{table}

\subsection{Hysteresis properties versus new Neural Networks} 
In the able Table~\ref{table2} all four types of neural networks have been compared with each other in terms of the learning capabilities of different properties of hysteresis behaviors. As it can be seen from Table ~\ref{table1} it can be concluded that the EPNN is more general than other neural networks and its capabilities of learning hysteresis behaviors with different characteristics is obvious.

\begin {table}[H]
\caption {Comparing all 4 types of neural networks } \label{table2}
\begin{center}
	\begin{tabular}{|l||c|c|c|c|}
		\hline Hysteresis properties& PNN & GPNN & Preisach-NN & EPNN \\ 
		\hline \hline Masing & Y & Y & Y & Y \\ 
		\hline Non-masing & N & N & Y & Y \\ 
		\hline Symmetric & Y & Y & Y & Y \\ 
		\hline Asymmetric & N & N & Y & Y \\ 
		\hline Rate-independent & Y & Y & Y & Y \\ 
		\hline Rate-dependent & N & N & N & Y \\ 
		\hline Congruent & Y & Y & Y & Y \\ 
		\hline Non-congruent & N & Y & N & Y \\ 
		\hline Algorithm & GA & GA & SGSD & HA \\ 
		\hline 
	\end{tabular} 
\end{center}
\end{table}

\subsection{Training algorithm}
The adaptive parameters of extended Preisach-NN are its connection weights and $\beta$-values  of the NDS neurons. Generally the error of the Preisach-NN can be expressed in terms of the adaptive parameters. The Mean Squared Erorr (MSE) of the extended Preisach-NN is 
\begin{equation}
\mathrm{MSE(\mathbf{w},\boldsymbol{\beta})}=\dfrac{1}{N}\sum\limits_{i=1}^{N}(\hat{y}(t_i)-y(t_i))^2 \label{MSE}
\end{equation}
where $\mathbf{w}$, $\boldsymbol{\beta}$, $N$, and $\hat{y}$  are connection weight vector, vector of the $\beta$-values of the NDS neurons, number of the training pairs $(x(t_i),y(t_i))$,  and  the output of the extended Preisach-NN, respectively. In Eq.~\ref{MSE} the output of the extended Preisach-NN is only related to the adaptive parameters $\mathbf{w}$ and $\boldsymbol{\beta}$. These parameters must be tuned through the training algorithm.

The backward propagation of errors cannot be used to calculate the gradient of the MSE with respect to all the weights ($\mathbf{w}$) because the activation function of NDS neurons is not differentiable. In addition there exists other adaptive parameters ($\boldsymbol{\beta}$) in the extended Preisach-NN for which the method of the backpropation cannot be applied. In this situations, the most effective algorithms which can be utilized to optimize the objective function (i.e. error function) are meta-heuristic algorithms. This group of algorithms are suitable for optimization problems in which the exact model is not available and due to the complexity of the solution space it is most likely that the algorithm can be trapped in a local optima. Because of the mentioned pros they are applied in most of the engineering fields such as scheduling \cite{jafarzadeh2017solving}, supply chain \cite{jafarzadeh2017genetic} and so on. In this work, Genetic Algorithm (GA) as a zeroth order optimization method has been used for the training of the Prandtl-NNs successfully \cite{joghataie2008dynamic,joghataie2011matrix,farrokh2015modeling}. Farrokh and Joghataie \cite{farrokh2013adaptive} successfully used the sub-gradient method with space dilatation \cite{shor2003algorithms} for training process of the Preisach-NN. The convergence rate would be slow if the GA was used in the traning of the Preisach-NN since it has lots of connection weights due to the second hidden layer and the sigmoid neurons. However the sub-gradient method like the other gradient methods is subject to be trapped in local minimum due to the inappropriate starting point.      

The algorithms used for the training of the Prandtl-NNs and the Presiach-NN have their props and cons. GA suffers from the convergence rate in case there are many free parameters and the sub-gradient method has the initial point problem. Fortunately the methods advantages compensate each other for the disadvantages. Usually finding a suitable first start point in the gradient-based algorithms is very difficult. However, GA is useful to find appropriate point as an initial point of the sub-gradient method  because of the randomness in its nature. Instead, the sub-gradient method can expedite the training process when the GA  cannot reduce the error any longer.

A hybrid algorithm based on the combination of an enhanced GA  \cite{jafarzadeh2017enhanced} and the sub-gradient method with space dilatation \cite{shor2003algorithms,shor1998subgradient} has been adopted for the training of the extended Preisach-NN in this paper. According to the hybrid algorithm firstly the free parameters of the extended Preisach-NN are tuned by GA till the MSE will not be significantly reduced. Afterwards using the best individual obtained by GA as an initial point of the sub-gradient method, the second phase of the training process starts. It should be noted that the activation function of the output neuron of the extended Preisach-NN is linear; therefore, the weights of its connections can be determined implicitly in terms of the other free parameters of the neural network through the least squares \cite{farrokh2013adaptive}. Using this property the number of the independent free parameters of the extended Preisach-NN shown in Fig.~\ref{EPNN} is $n_{\mathrm{stop}}\times n_{\mathrm{tanh}}+2(n_{\mathrm{stop}}+n_{\mathrm{tanh}}+1)$ where $n_{\mathrm{stop}} $ and $n_{\mathrm{stop}} $ number of NDS neurons in first and number of tanh neuron in the second hidden layer, respectively. 


\section{Extended preisach neural network (XPNN)} \label{sec:Assessment}

For investigating the performance of the introduced model, real physical experimental results have been utilized. In order to assess the proposed model, we have selected different hystereses from different fields of engineering with different properties. We have used experimental data sets from different researchers in the field. It should be noted that for these data sets, researchers utilized different modeling methods to identify the hysteretic behavior. However, we apply the proposed model on them identically to show its generality. In addition the results of the EPNN are compared with the results of the other proposed hysteresis models in cases they were available.

\subsection{Example 1: Symmetric non-masing rate-independent, saturated}  
\textbf{Ferromagnetic material:}

\begin{figure}[ht]
	\centering
	\includegraphics [width=3in]{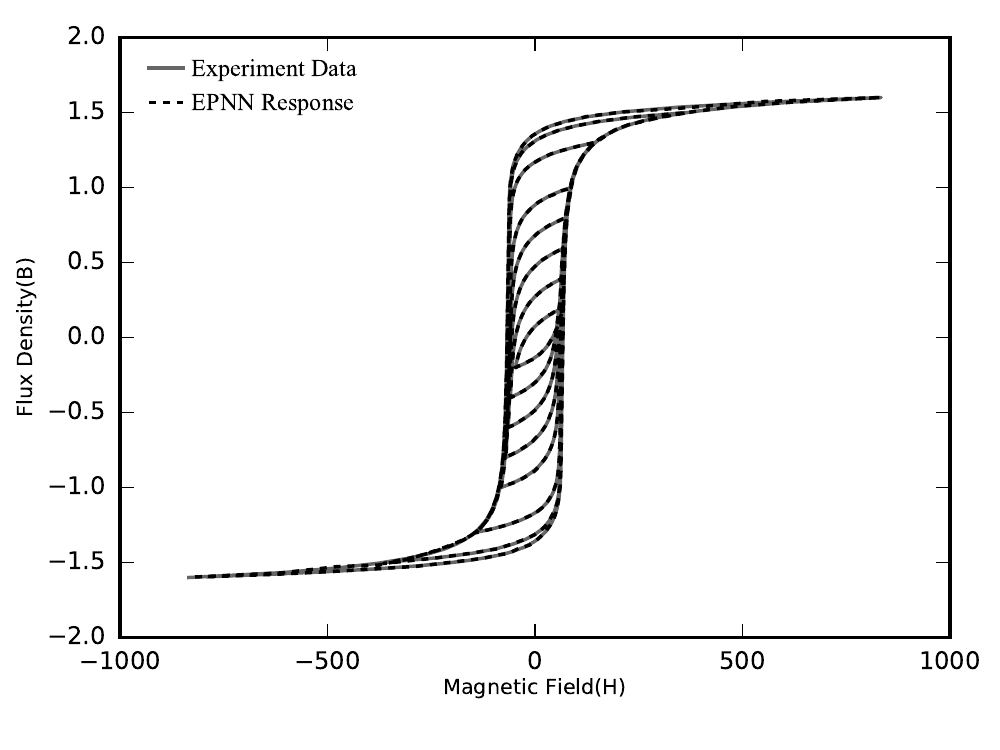}
	\caption{Hysteretic Relation in Ferromagnetic Material}\label{ex1}
\end{figure}

Fig.~\ref{ex1} illustrates the measured hysteretic relation between the applied magnetic field and the response flux density of a ferromagnetic material. The reported results have shown very similar trends in view of the hysteresis phenomenon [1, 82], which are summarized below: As it can be seen from Fig.~\ref{ex1}, hysteresis loops are not follows masing rules. Furthermore, hysteresis loops are symmetric and saturated. The output flux density (B) tends to saturate as the input field (H) exceeds certain limit that may depend upon the properties of the material. The hysteresis loops are generally considered rate-independent and show insignificant variations under inputs in the low frequency range. The proposed model, EPNN, has been applied to learn the hysteresis behavior. The results show that EPNN has capability of learning symmetric non-masing hysteresis loops very precisely.

\subsection{Example 2: Non-congruent non-masing}  
As it can be seen schematically, in the First example hysteresis loops are non-congruent, non-masing and rate-independent. Since PNNs only can learn congruent masing hysteresis behaviors which in turn is because of intrinsic property of stop neurons, therefore, they are not suitable neural networks to use in learning non-congruent hysteresis behaviors. Although, at some point, GPNN can be applied in learning non-congruent non-masing hysteresis, as a whole, it has not comprehensive capability of learning these behaviors smoothly and precisely. Hence, generally it can not be assumed a appropriate model to learn non-congruent non-masing hysteresis behaviors. Finally, Preisach-NNs are another type of neural networks which are designed to learn non-masing hysteresis behaviors as well as masing ones. However, owing to the fact that it is emanated from preisach model which merely can create conguent hysteresis loops, so that this type of neural networks can non be applied in learning non-congruent hysteresis behaviors. According to above explanations none of mentioned neural networks are as much suitable as they should be to apply in learning non-congruent non-masing hysteresis behavior of first example.   

\begin{figure}[ht]
	\centering
	{
		\includegraphics [width=3in]{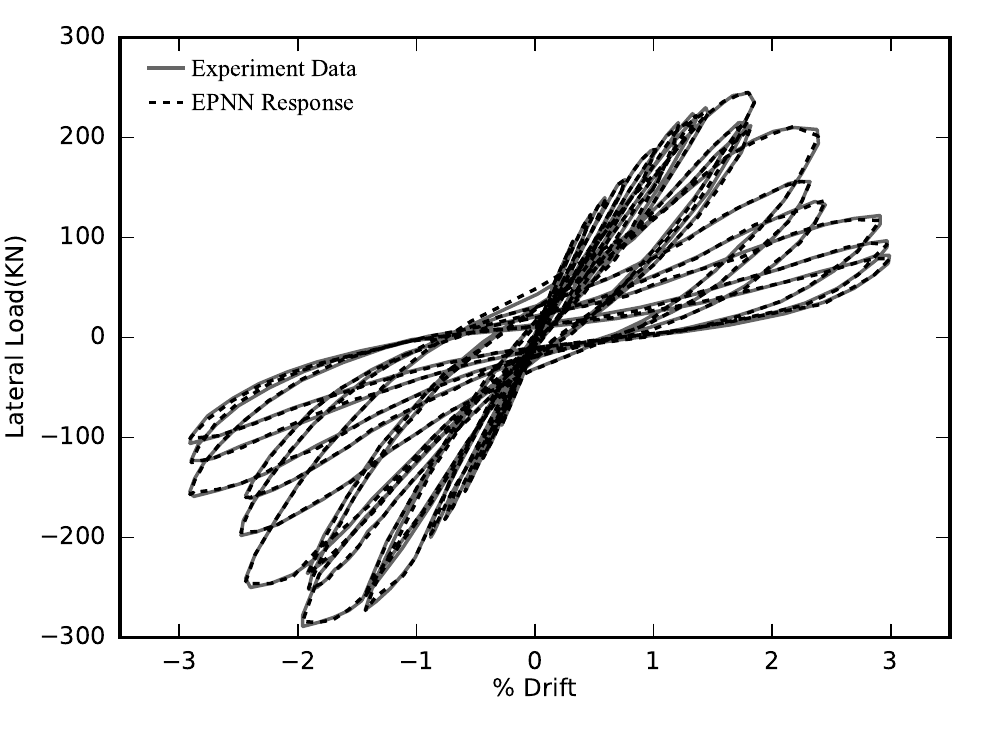}
	}
	{
		\includegraphics [width=3in]{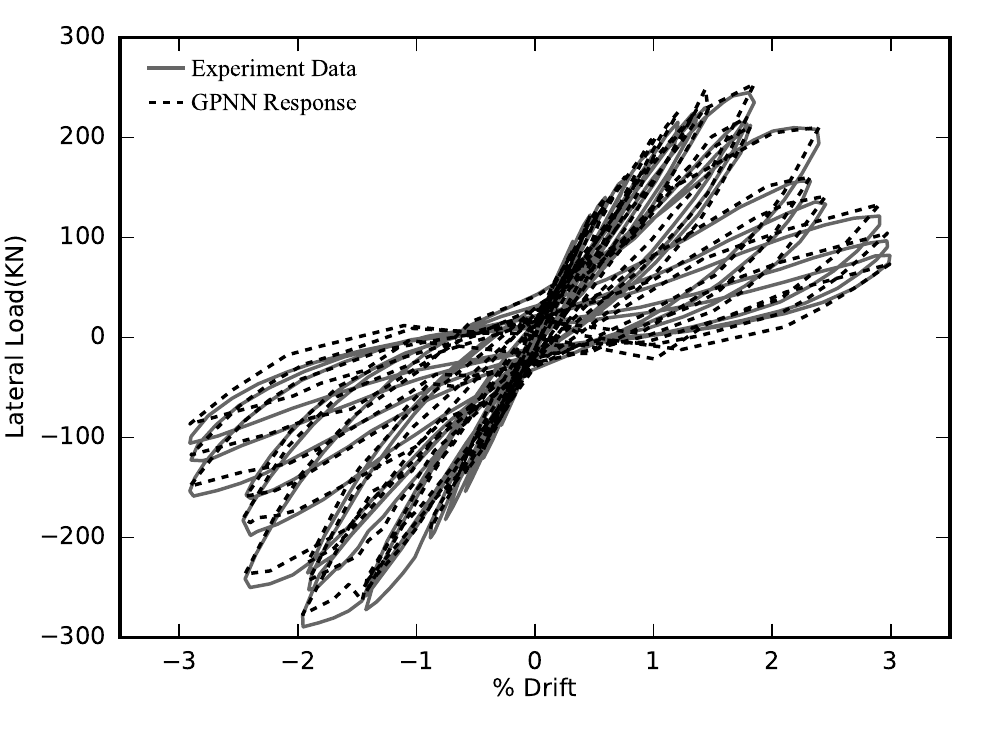}
	}
	\caption{Hysteresis Loop. Hysteresis obtained from EPNN (Top), Hysteresis obtained from GPNN (Bottom)}\label{ex2_2}
\end{figure}

We utilize the results of the experiments which Clyde et al. (2000) \cite{clyde2000performance} conducted on exterior beam–column joints of plane multi-story concrete frames. In their experiments, they measured x = drift ratio versus y = the lateral load of the frames. The hysteretic behavior is mainly subjected to damage; therefore, most of the Preisach-type models and Prandtl models are not useful for them because damage index in hysteresis loops is increased even if the loops have the identical extremum values and it means the loops are non congruent.  Fig.~\ref{ex2_2} shows the time histories of drift ratio and lateral load (solid gray curve), respectively. Extended Preisach Neural networks has been utilized to learn the hysteresis behavior. The architecture of the EPNN is described in the previous sections. It should be noticed that training of the EPNN has not been done on the whole interval of the time histories ?? and the remaining has been set aside for test. The intervals used for training and test have been shown in Figure 6(a)?.

\begin{figure}[ht]
	\centering
	\includegraphics [width=3in]{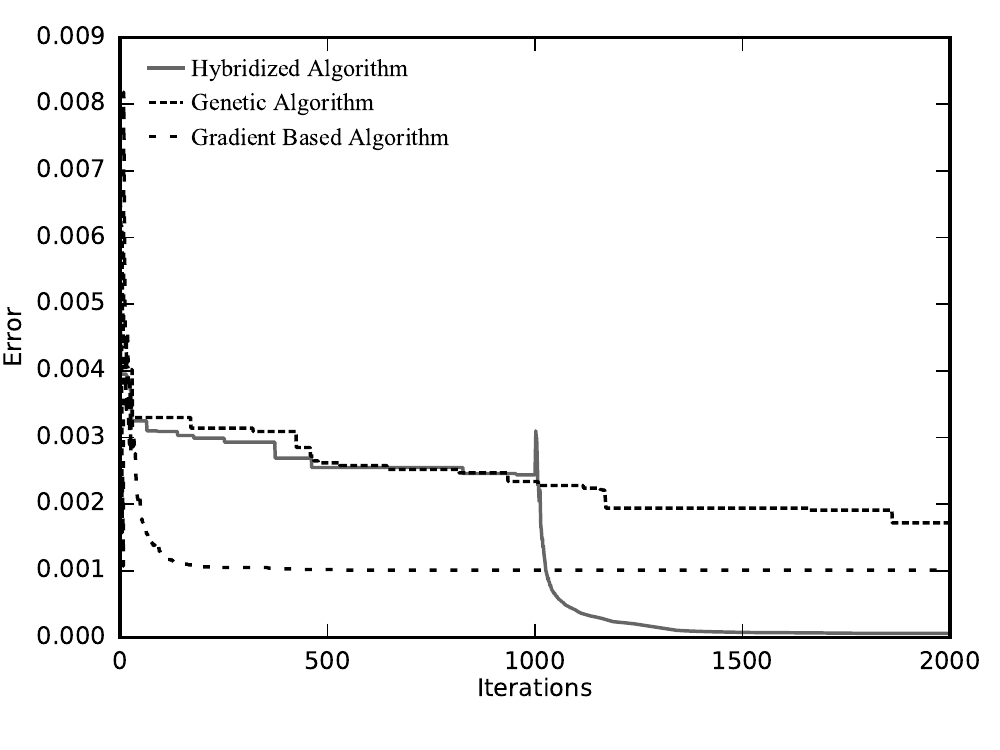}
	\caption{Training Convergence}\label{ex2_e}
\end{figure}

Using the novel proposed algorithm known as hybridized training algorithm which is combination of Genetic Algorithm and gradient-based algorithm, training has been performed for 2000 epochs. It should be noted that the training process has two step: 1) training using GA and 2) training using gradient based algorithm. Owing to that, first, the training process is continued 1000 epochs using GA and then the training process for the rest of epochs is pursued using gradient based method. It should be noticed that the first start point of gradient-based algorithm is the last optimal point obtained from GA. The training convergence of the EPNN has been shown in Fig.~\ref{ex2_e}

In the first step of training process the Mean Square Error (MSE) of EPNN has been reduced from 41 to 0.004 during training. In the second step of training process the Mean Square Error (MSE) of EPNN has been reduced from 0.004 to 0.0001 during training process. In order to show the superiority of the new proposed algorithm the training process is conducted for three separated cases for 2000 epochs alone: 1) Only GA is used for training, 2) Only gradient based algorithm is utilized for training and 3) The combination of both which is called HTA is used for training process. As it can be seen in Fig.~\ref{ex2_e}, using the new proposed algorithm the error of EPNN has been decreased much more than the aforementioned algorithms when be applied alone. Therefore it can be concluded that in comparison with GA and gradient based algorithm, the hybridized training algorithm is much more successful in training process of EPNN. After the completion of the training, the EPNN has been utilized to simulate for training interval and to predict for test interval of the considered hysteretic behavior. Solid curve in Fig.~\ref{ex2_e} shows the time history of the relative error of the trained EPNN for both simulation and prediction. The simulation errors are very low and the prediction errors is limited to 3.2

In order to show the performance of EPNN, the comparison between the EPNN response and the other model is helpful. Recently Farrokh et al. (2015) \cite{farrokh2015modeling} proposed a new neural network for modeling deteriorating hysteretic behavior and called it Generalized Prandtl Neural Network (GPNN). They also applied their proposal on this case. Their results for this case have been summarized in Figures 6(b) ? and (c). ? The comparison of the absolute relative errors of GPNN and EPNN, depicted in Figures 6(c) ? and (d) ? respectively, shows EPNN can simulate the hysteretic behavior with much higher precision than GPNN. Also EPNN has a better performance than GPNN in prediction.

The simulation and perdition MSE of GPNN for this case are 99.47 and 140.13, respectively. Comparison of these values to the MSEs of EPNN show that EPNN is more precise than GPNN in this case. In Fig.~\ref{ex2_2}(a) and .~\ref{ex2_2}(b) hysteresis obtained from EPNN and GPNN respectively have been compared with experimental data. It is obvious that EPNN follows each loop of the considered hysteretic behavior accurately. In contrast the difference between GPNN response and experimental data is observable in Fig.~\ref{ex2_2}(b).

\subsection{Example 3: Asymmetric non-masing}  
\begin{figure}[ht]
	\centering
	{
		\includegraphics [width=3in]{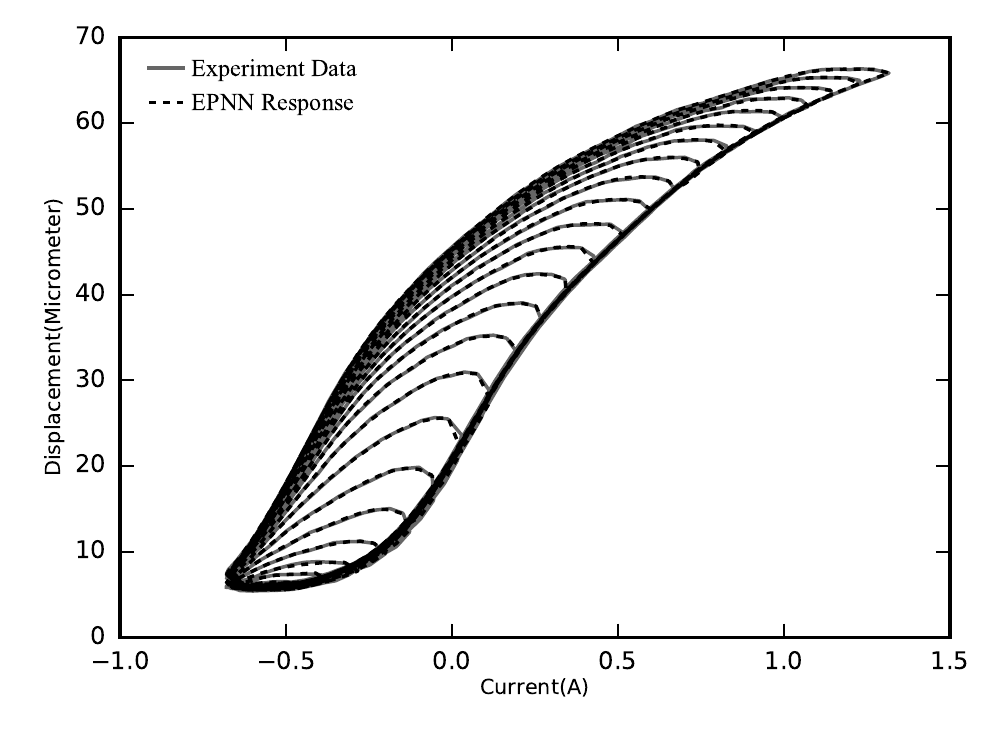}
	}
{
		\includegraphics [width=3in]{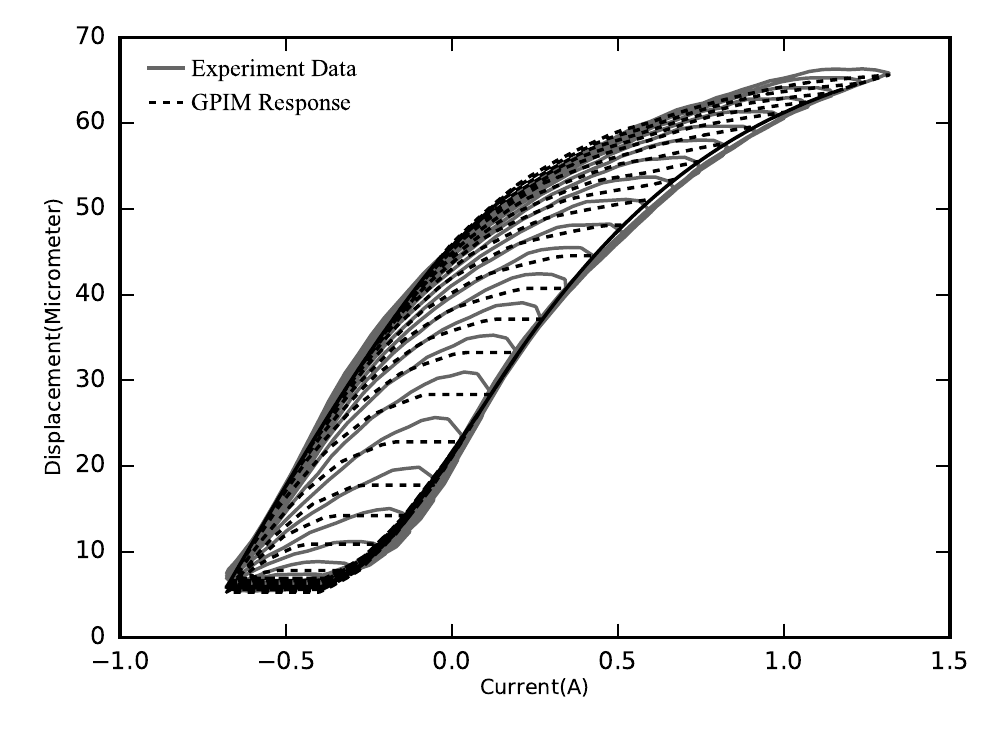}
	}
	\caption{The measured output displacement versus input current relationships for the magnetostrictive actuator. Hysteresis obtained from EPNN (Top), Hysteresis obtained from GPNN (Bottom)}\label{ex3}
\end{figure}

Magnetostriction is a phenomenon of strong coupling between the magnetic and mechanical properties of some ferromagnetic materials. Magnetostrictives the same as piezoelectrics, and shape memory alloys which are considered as smart materials have built-in sensing/actuation capabilities. Tan and Baras (2004, 2005) \cite{tan2004modeling,tan2005adaptive} conducted experiments on a magnetostrictive (Terfenol–D) actuator. The actuator response was quite rate-independent when it operated in a low frequency range (typically below 5Hz). For this phase, they investigated the major and minor hysteresis loop properties \cite{tan2005adaptive}. They tested the considered magnetostrictive actuator under increasing triangular input currents with amplitude ranging from -0.7 to 1.2 A. The measured output displacement versus input current relationships for the magnetostrictive actuator, shown as solid gray curves in Fig.~\ref{ex3}, has been adopted for the assessment of our proposal. The measured data obviously displays the asymmetric non-masing input-output relationships, output saturation and asymmetric major and minor hysteresis loops for the magnetostrictive actuator. 

\begin{figure}[ht]
	\centering
	\includegraphics [width=3.in]{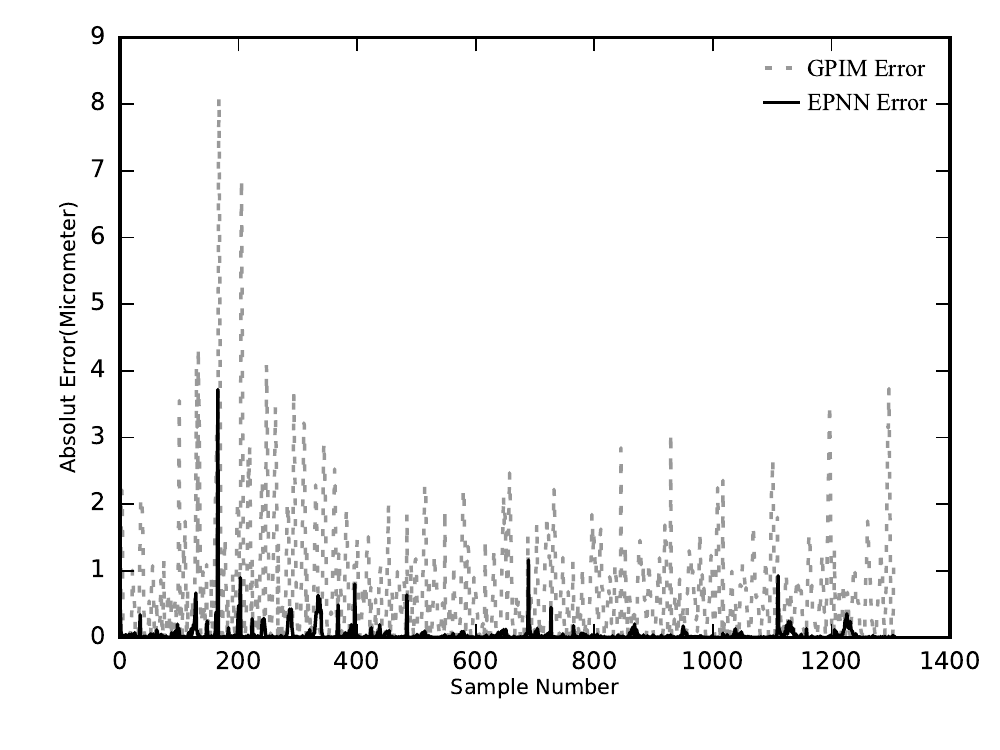}
	\caption{Compare the relative error of these two approaches}\label{ex3_e}
\end{figure}

The same as previous example, EPNN with the same architecture has been considered for this example. The validity of the EPNN is investigated by comparing the EPNN response with the available measured data of the magnetostrictive actuator. After its training, it has been used for the simulation of the considered hyeteretic behavior in this case. Fig.~\ref{ex3}~(a) illustrate comparisons of the EPNN response of the magnetostrictive actuator, with the measured data. The results clearly suggest that the EPNN can effectively predict the asymmetric non-masing hysteresis properties of magnetostrictive actuator. In order to verify the superiority of our proposed model with other models, the result of EPNN is compared with Generalized Prandtl–Ishlinskii Model proposed by Al Janaideh et al. (2011) \cite{al2010analytical} for simulation of hysteretic bahaviors of smart actuator. They utilized their proposal on the hysteresis of this example. In Fig.~\ref{ex3}~(a)~and~(b) the responses of the EPNN and the GPIM have been compared with the original data, respectively. The better performance of the EPNN to the GPIM is observable because EPNN follows the hysteresis loop more precisely than GPIM. In addition, Fig.~\ref{ex3_e} compares the relative error of these two approaches. The effectiveness of the EPNN in predicting the asymmetric non-masing saturated hysteresis response of the magnetostrictive actuator can also be seen from comparisons of the displacement responses in the time domain, presented in Fig.~\ref{ex3_e}. The relative error of the EPNN is lower than the GPIM for each sample. The simulation MSEs for EPNN and GPIM are 0.016 and 0.654, respectively.

\subsection{Example 4: Asymmetric output-saturated} 

The output-input hysteresis properties of two shape memory alloy (SMA) actuators have been measured by Robert Gorbet \cite{tan2005adaptive}. SMA materials, used as wires or bars, have many applications in engineering especially in mitigation systems \cite{Farzad,FarzadS}. The study performed measurements on two SMA actuators, including a one-wire and two-wire actuators. The measurements were performed to establish relationships between the output displacement and the input temperature. The variations in input temperature were realized by applying triangular waveform currents of varying magnitudes. In this study only the data from two-wire SMA actuator has been used in order to test the proposed model. The input current for the two-wire SMA actuator ranged from -1 to 1 A resulting temperature variations from -175 to 175° C. 
\begin{figure}[ht]
	\centering
	\includegraphics [width=3in]{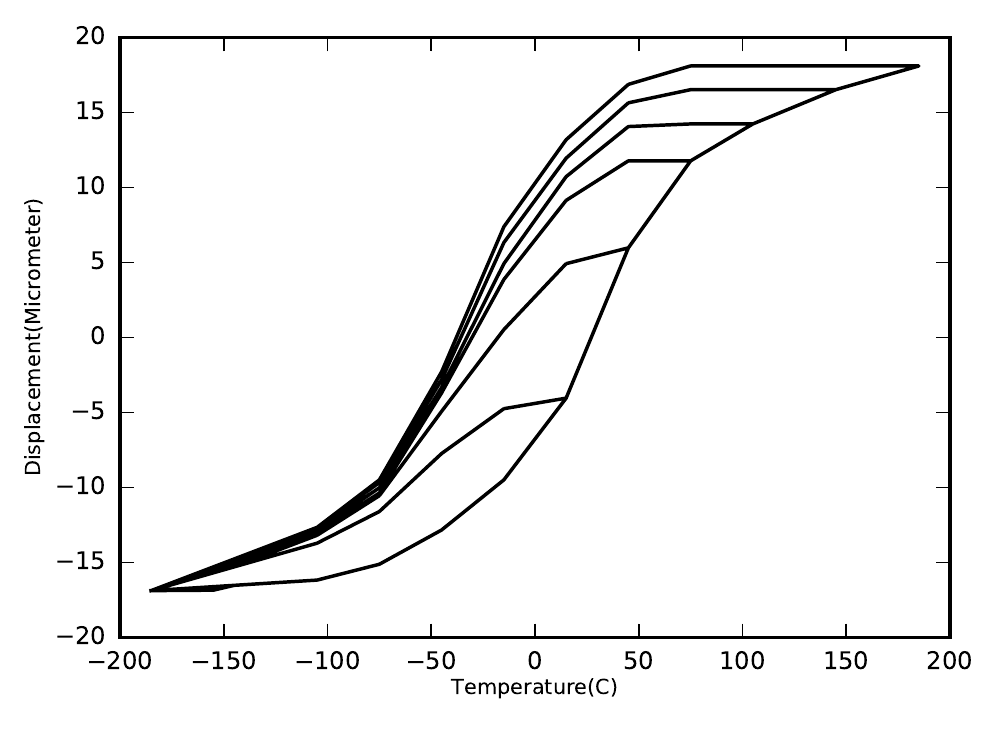}
	\caption{The output-input in SMA actuators}\label{ex4_io}
\end{figure}
\begin{figure}[ht]
	\centering
{
		\includegraphics [width=3in]{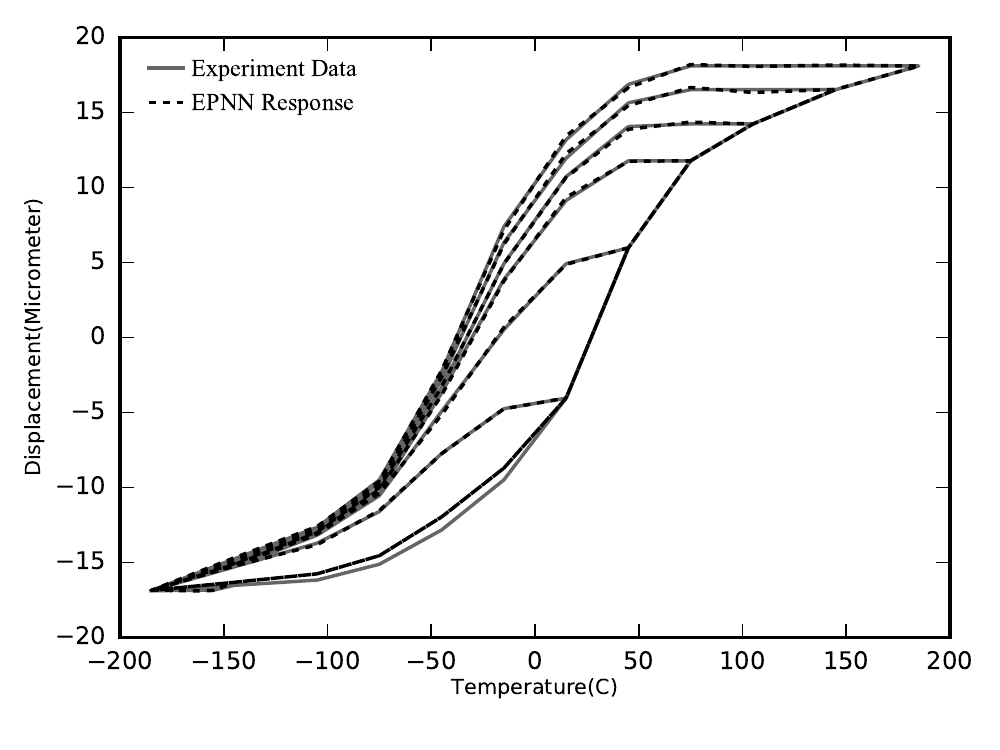}
	}
	{
		\includegraphics [width=3in]{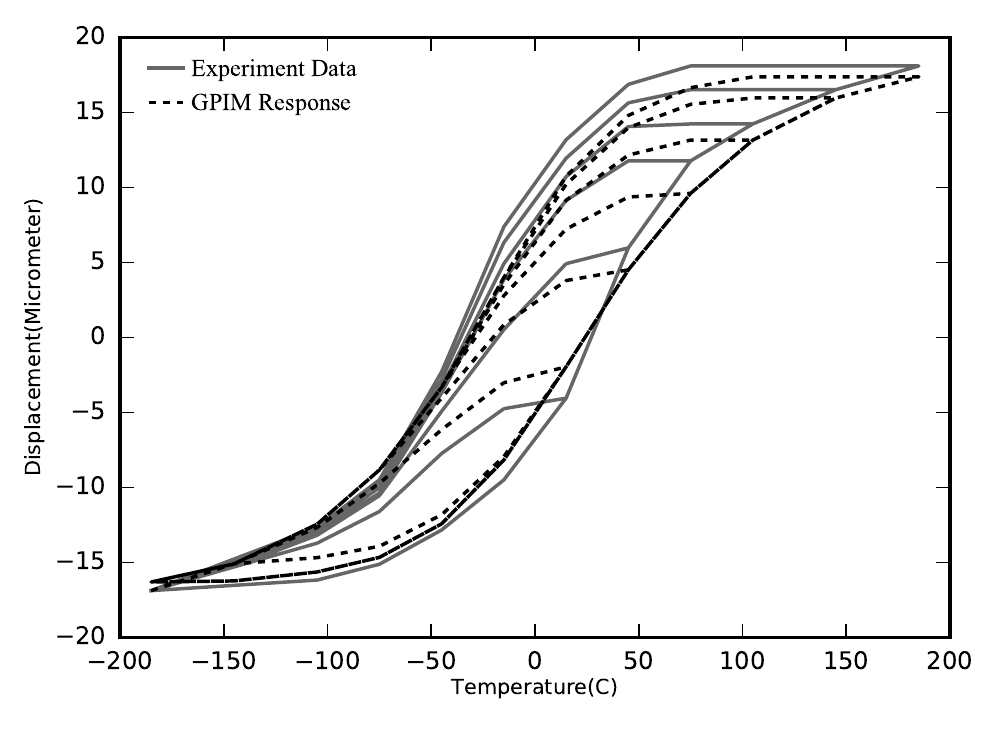}
	}
	\caption{Responses of the EPNN and the GPNN. Hysteresis obtained from EPNN (Top), Hysteresis obtained from GPNN (Bottom)}\label{ex4}
\end{figure}
The input current variations were at a low frequency. The acquired data were analyzed to derive the minor and major hysteresis loops of the two SMA actuators, while the rate dependence of the actuators output and hysteresis could not be established. The output displacement responses of the actuator are illustrated in Fig.~\ref{ex4_io} as a function of variations in the input temperature. The results clearly show highly asymmetric output-input relations of actuator. The asymmetry is evident in both the major as well as minor hysteresis loops. The results also show notable output saturation. The two-wire actuator exhibit saturation when the magnitude of the temperature approaches 50°C.

EPNN with the same architecture has been considered for this example Fig.~\ref{ex4} illustrates comparisons of the major and minor loops in the displacement responses of the model with the corresponding measured data. The results suggest reasonably good agreements between the model and measured displacement responses. After its training, it is used for the simulation of the considered hyeteretic behavior in this case. In order to verify the superiority of our proposed model with other models, the result of EPNN is compared with Generalized Prandtl–Ishlinskii Model proposed by Al Janaideh et al. (2011) \cite{al2010analytical} for simulation of hysteretic bahaviors of smart actuator. They utilized their proposal on the hysteresis of this example. In Fig.~\ref{ex4}~(a)~and~(b) the responses of the EPNN and the GPIM have been compared with the original data, respectively. The better performance of the EPNN to the GPIM is observable because EPNN follows the hysteresis loop more precisely than GPIM.

\subsection{Example 5: Asymmetric non-masing non-congruent}  
As the fifth example, we use the results of the experiments which Sinha et al.~\cite{sinha1964stress} did on exploring the cyclic un-axial compression response of plain concrete. Important work for calibrating a hysteretic compression model for concrete. They researched on Stress-strain relations for concrete under cyclic loading. In their experiments, they measured x = strain y = stress. The hysteretic behavior is mainly subjected to damage; therefore, most of the Preisach-type models are not useful for them because damage index in hysteresis loops is increased even if the loops have the identical extremum values and it means the loops are not congruent. Even though GPNN are capable of learning non-congruent hysteresis behaviors, as it can be seen, hysteresis behavior of the stress-strain relations for concrete under cyclic loading are asymmetric, non-masing and non-congruent; therefore PNN,GPNN and Preisach-NN are not capable of learning this type of hysteresis behavior precisely.

\begin{figure}[H]
	\centering
	\includegraphics [width=2.5in]{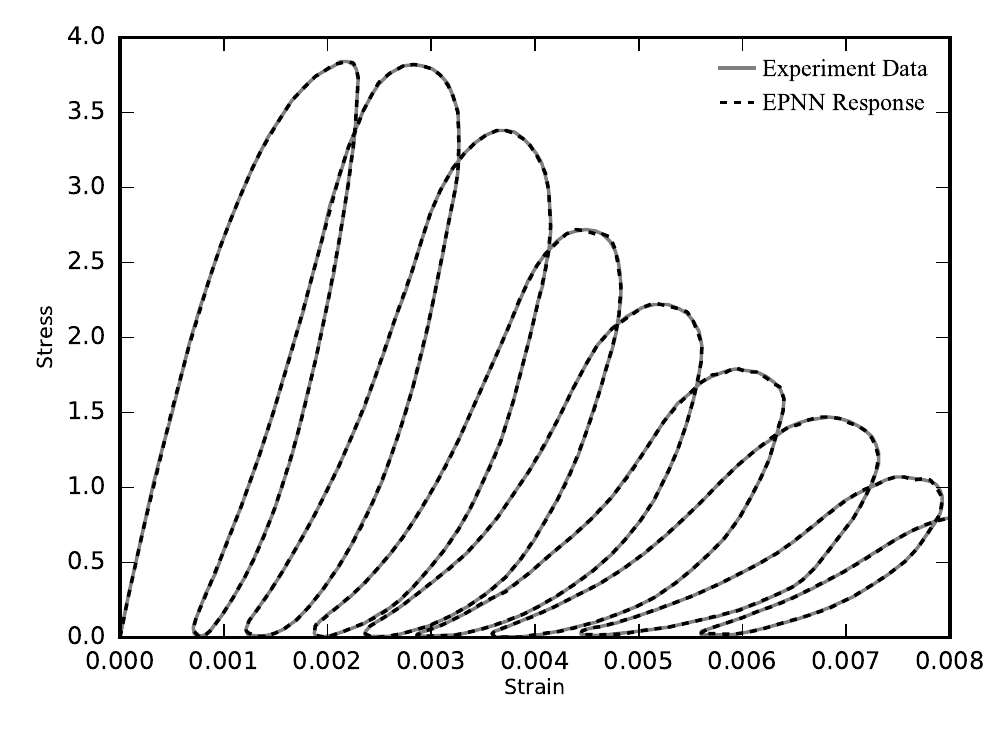}
	\caption{Experimental hysteresis loops and the simulated loops}\label{ex5}
\end{figure}

EPNN with the same architecture has been considered for this example as well. Fig.~\ref{ex5} shows both experimental hysteresis loops and the simulated loops by proposed model in this paper. It has capability of learning the case quite precisely with Mean Square Error (MSE) 2.79. The validity of the EPNN in predicting the asymmetric, non-masing as well as non-congruent major as well as minor input–output hysteresis loops was demonstrated on the basis of the available measured data.

\subsection{Example 6: Symmetric rate-dependent}  

Piezoelectric translator (PZT) actuators are increasingly popular for high-speed and high-accuracy micro- and non-opositioning systems because of its high output force, large bandwidth and fast response time \cite{gu2011modeling}. However, the hysteresis of the PZT actuators is a rate-dependent phenomenon, strongly depending on frequencies and amplitudes of the control input. Therefore, the Preisach and Prandtl-Ishlinskii models have not capability for the applications with high-speed input. To represent the hysteresis effects, experiments are performed with designed sinusoidal excitations under different frequencies in the range 0.5–300 Hz by GuoYing Gu and LiMin Zhu in 2010~\cite{gu2011modeling}. The reported data for various piezoelectric actuators under excitation of varying magnitudes and frequencies suggest nearly rate dependent symmetric major and minor hysteresis loops. 

\begin{figure}[H]
	\centering
	\includegraphics [width=2.4in]{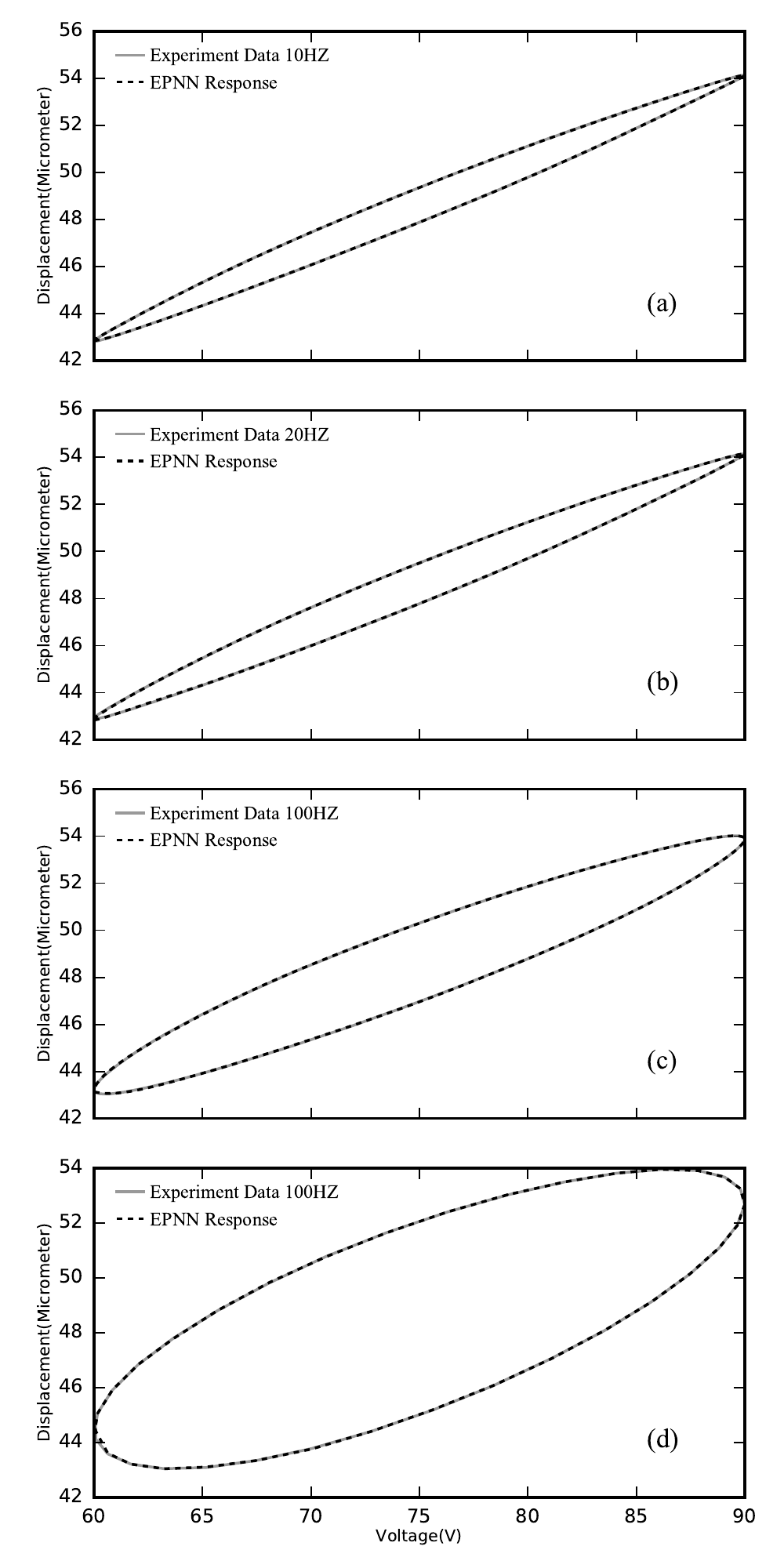}
	\caption{Output displacement of the actuator versus input voltage for different frequencies}\label{ex6_freq}
\end{figure}

\begin{figure}[H]
	\centering
	\includegraphics[width=2.5in] {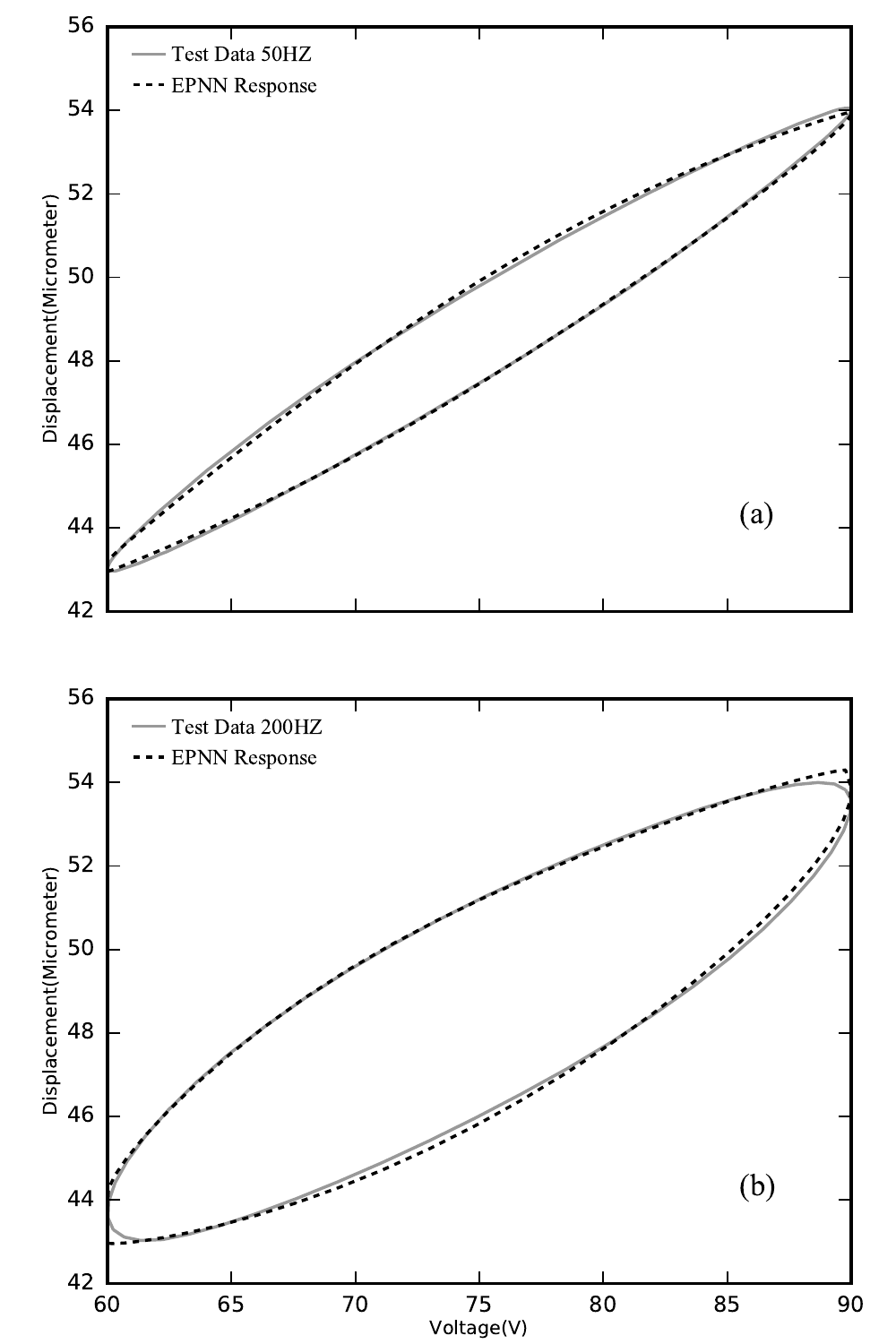}
	\caption{The model generality with not training data}\label{ex6_freq2}
\end{figure}

In this paper the aforementioned experimental data is applied to validate the EPNN. Since the hysteresis is rate-dependent, $\stackrel{.}{x}$ has been considered in the input layer of the EPNN. Fig.~\ref{ex6_freq} shows measured output displacement of the actuator versus input voltage for different frequencies as solid curves. In this figure the estimated output displacements of the actuator by EPNN has been depicted versus input voltages by dashed curves. Again the good performance of the EPNN in simulation is observable. To show the generality of the proposed model, two frequencies 50Hz and 200Hz is selected to test the model (Fig.~\ref{ex6_freq2}). The data regarding to these frequencies are not in the training data. The results show the good learning capability of the EPNN for not only training data but also for the testing data which never were in the training data.

\begin{figure}[ht]
	\centering{
		\includegraphics [width=2.5in]{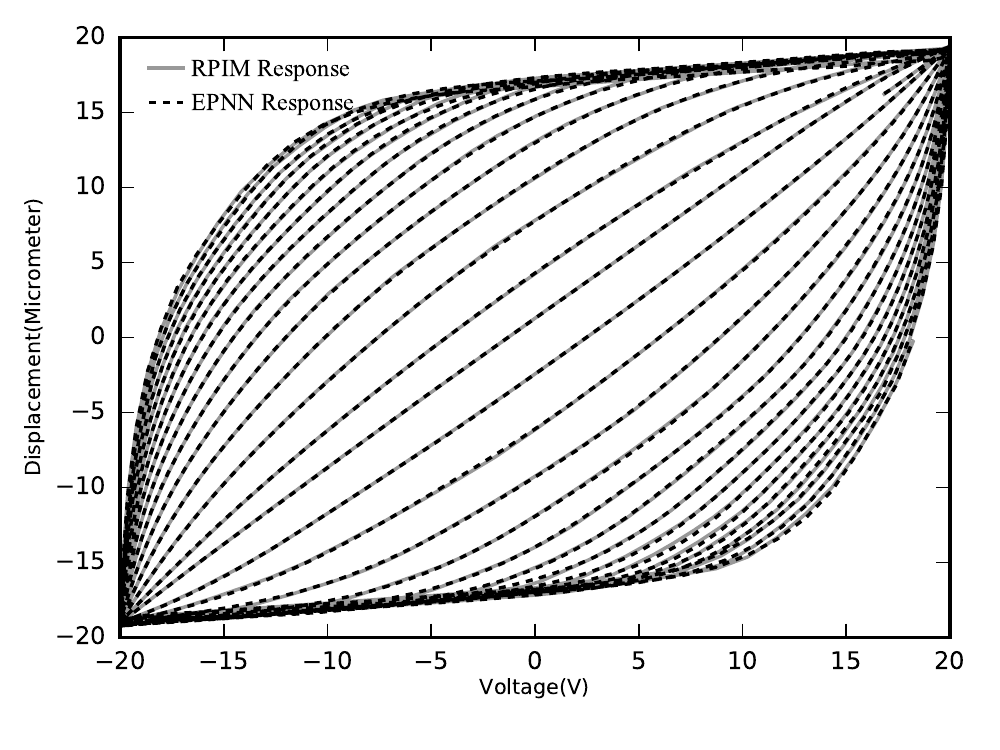}}{
		\includegraphics [width=2.5in]{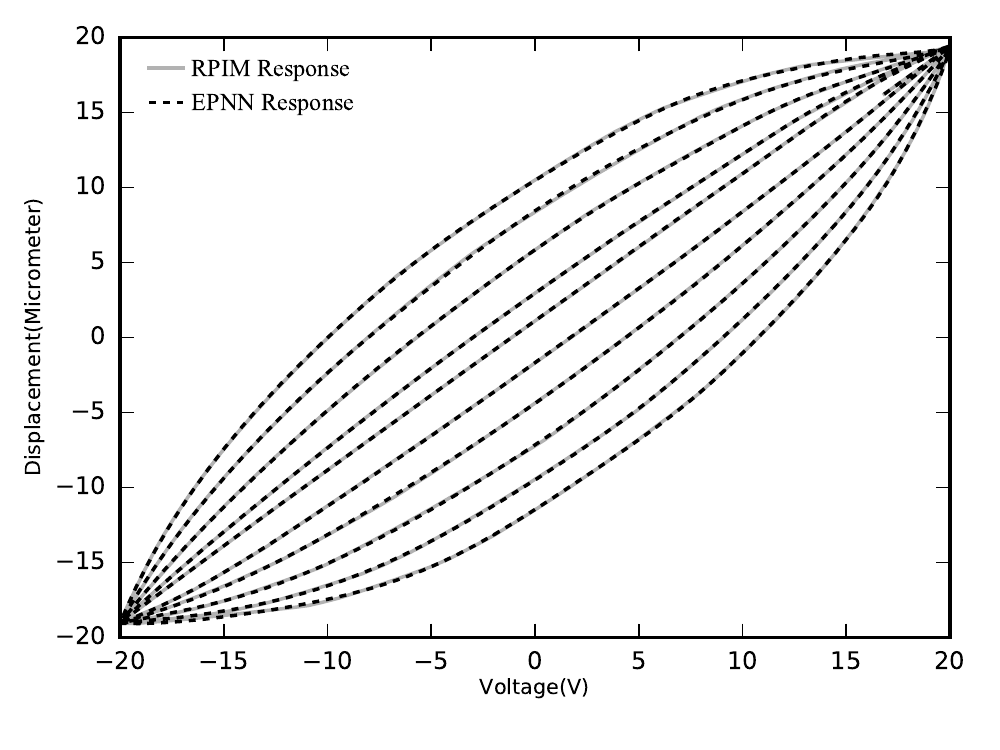}
	}
	\caption{Rate-dependent hysteretic behavior of the smart actuators. Rate-dependent hysteresis for training algorithm (Top), Rate-dependent hysteresis for testing algorithm (Bottom)}\label{ex6_aljonaide}
\end{figure}

We have used for a hysteresis model for generating some data on a symmetric rate-dependent hysteresis for both training and test of the EPNN. Al Janaideh et al.~\cite{al2012inverse,al2011inversion,al2010analytical} proposed a novel approach in order to simulate rate-dependent hysteresis. They suggested a Rate-dependent Prandtl-Ishlinskii Model (RPIM) to describing the rate-dependent hysteretic behavior of the smart actuators. Using their model two different sets of data are synthetically generated for the assessment of the proposed EPNN on rate-dependent hysteresis. The response of the RPIM for input signals shown in Fig.~\ref{ex6_aljonaide}~(a)~and~(b) used respectively for training and test of the considered EPNN for this case. The hysteresis loops of the generated data set have been shown in Fig.~\ref{ex6_aljonaide} as solid gray curves. To investigate the rate dependence, we have tuned two different EPNNs with and without $\stackrel{.}{x}$ in the input vector on the data set. The simulation errors for them in terms of MSE were 778 for the EPNN without $\stackrel{.}{x}$ and ${7}\times{10}^{-5}$ for the EPNN with $\stackrel{.}{x}$.

The obtained results show that the data set is rate-dependent and   $\stackrel{.}{x}$ cannot be ignored in this case. The testing the EPNN shows the very good performance because the MSE value on test data is 0.002 and very low. It should be noted that the frequency contents of the test data is different from the training data. However the responses of the NMM are not distinguishable from the RPIM responses in Fig.~\ref{ex6_aljonaide} for both training and test data. Results comparison show the excellent performance of the EPNN for this case.
Comparisons of the model results with the measured data for a PZT actuator revealed reasonably good agreements.

\subsection{Example 7: Asymmetric rate-dependent non-congruent}  
Magnetostriction is the phenomenon of strong coupling between the magnetic and mechanical properties. Some ferromagnetic materials as Terfenol-D show this phenomenon between the output strain and the applied magnetic field. The output strains are produced due to the applied magnetic field which produces changes in magnetization. Magnetostrictive actuators have been widely used in micro-positioning applications and vibration control.
The major and minor hysteresis loop properties of a magnetostrictive actuator have been measured by Xiaobo Tan~\cite{tan2004modeling,tan2005adaptive}.
Various magentostrictive actuators show dependency of the output displacement on the rate of change of input current. It has been shown that the hysteresis increases under higher frequency excitations. On the basis of laboratory measurements, it has been shown that hysteresis in magentostrictive actuators is highly asymmetric, non-congruent and the output displacement is strongly rate-dependent beyond certain frequencies. Tan and Baras~\cite{tan2004modeling} performed laboratory measurements to characterize the hysteresis properties of a magentostrictive actuator under sinusoidal input currents of constant amplitude (0.8 A) with a bias of 0.1 A in the 10-300 Hz frequency range. The reported data revealed rate-dependent and asymmetric non-congruent hysteresis loops between the input current and the output displacement, particularly under excitations above 10 Hz. In a similar manner, the hysteresis characteristics of a magentostrictive actuator measured under excitation in the 100-500 Hz range showed increase in the hysteresis with increasing excitation frequency.
\begin{figure}[ht]
	\centering
	\includegraphics[width=\linewidth]{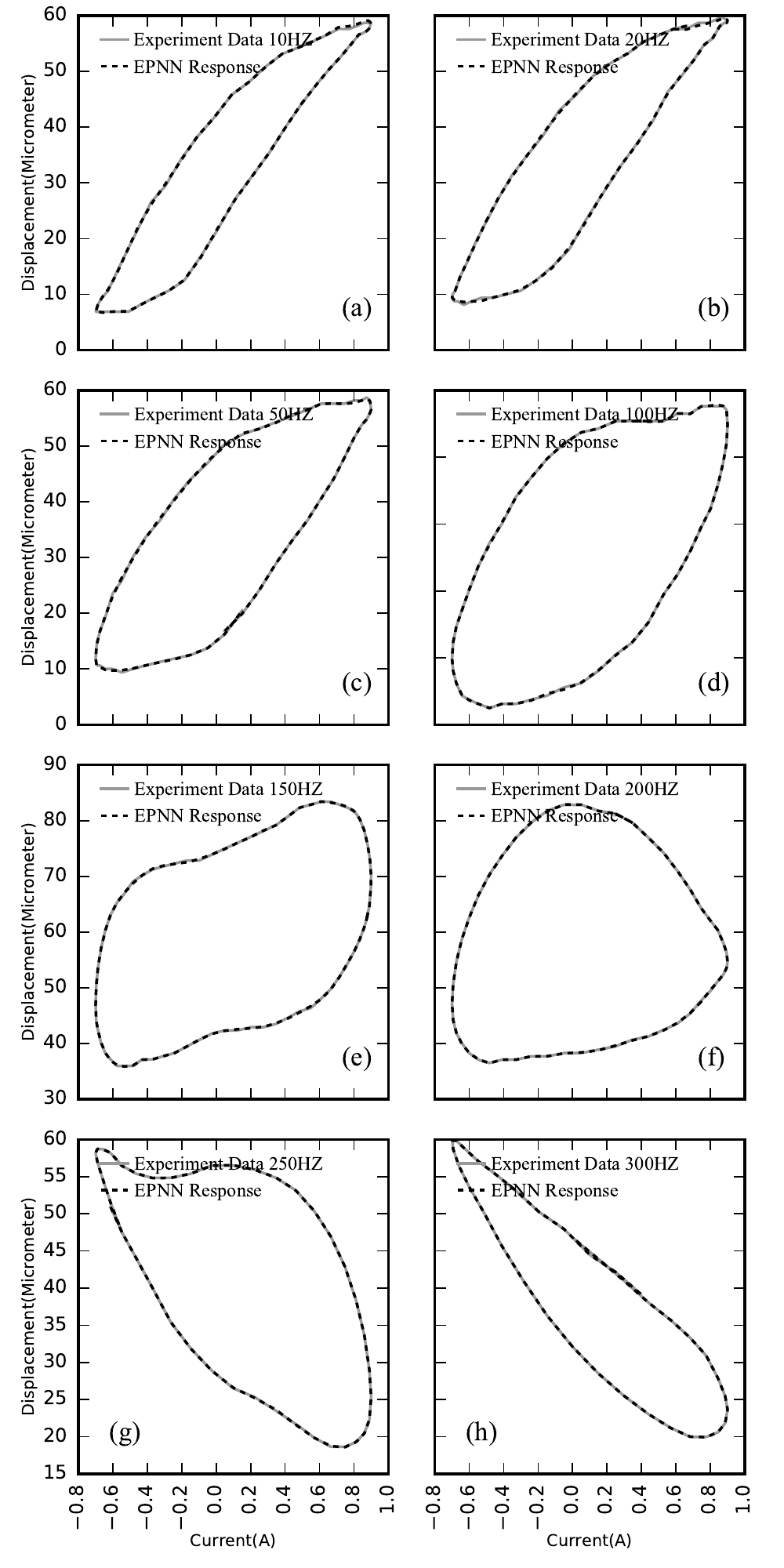}
	\caption{Hysteresis properties of a magnetostrictive actuator at different excitation frequencies}\label{ex7_freq}
\end{figure}
Fig.~\ref{ex7_freq} illustrates hysteresis properties of a magnetostrictive actuator at different excitation frequencies. The results attained under inputs at eight frequencies in the 10 to 300 Hz range show considerable variations in the hysteresis loops. As it can be seen in Fig.~\ref{ex7_freq} hysteresis loops of magnetostrictive actuators are asymmetric, non-congruent and rate-dependent. Therefore, PNN, GPNN and Preisach-NN are not capable of learning their behaviors. 

We used EPNN to learn rate dependent behavior of magnetostrictive. The proposed model is not only capable of learning asymmetric hysteresis behaviors, but also it can distinguish hysteresis loops whose their behaviors are rate-dependent.

The simulated hysteresis loops by EPNN have been depicted in Fig.~\ref{ex7_freq} as dashed curves which are not distinguishable from solid gray curves (experimental data) and this shows the very low simulation errors for this case too. Due to the lack of experimental data, in this case the EPNN test has not been performed.

\section{Conclusion} \label{sec:Conclusion}

In this paper, authors have improved previously proposed Preisach-NN and utilized it in the adaptive identification of the highly nonlinear hysteresis and it is called Extended Preisach neural network (EPNN). The extended Preisach-NN has two hidden layers; the first layer contains a bias to give capability of learning asymmetric hysteresis as well as symmetric ones, a linear and several DS neurons. The second layer has only sigmoidal neurons. In EPNN the only free parameters to be determined during its training are the connection weights between the neurons and β's which have been defined inside the DS neurons. In order to give EPNN the capability of learning non-congruent hysteresis behaviors, DS neurons are utilized in first hidden layer. Also, EPNN has obtained the capacity of learning rate-dependent hysteresis behaviors by putting the rate of changing inputs, x ̇(t), in the input layer. Hence, the proposed approach has capability of the simulation of the both rate independent and rate dependent hysteresis with either congruent or non-congruent loops as well as symmetric and asymmetric loops. The generality of the proposed model has been evaluated by applying it on various hystereses from different areas of engineering with different characteristics. The results show that EPNN is successful in the adaptive identification of the considered hystereses. The proposed neural network shows excellent agreement with experimental data from different areas of engineering.

\bibliographystyle{IEEEtran}
\bibliography{ref.bib}

\end{document}